%% file: main.tex
\documentclass[sigconf]{acmart}
\renewcommand\footnotetextcopyrightpermission[1]{} 
\acmDOI{}
\usepackage{booktabs} 
\usepackage{multirow}
\usepackage{subfig}
\usepackage{comment}
\usepackage{caption}
\usepackage{fancybox, graphicx}
\usepackage{tabularx}
\usepackage[export]{adjustbox}
\usepackage{makecell}
\usepackage{array}
\usepackage{balance}
\newcommand{\PreserveBackslash}[1]{\let\temp=\\#1\let\\=\temp}
\newcolumntype{C}[1]{>{\PreserveBackslash\centering}p{#1}}
\newcolumntype{R}[1]{>{\PreserveBackslash\raggedleft}p{#1}}
\newcolumntype{L}[1]{>{\PreserveBackslash\raggedright}p{#1}}

\def\BibTeX{{\rm B\kern-.05em{\sc i\kern-.025em b}\kern-.08emT\kern-.1667em\lower.7ex\hbox{E}\kern-.125emX}}

\begin{document}
\title{Knowledge Tracing with Sequential Key-Value Memory Networks}

\author{Ghodai Abdelrahman and Qing Wang\\} 
  \affiliation{%
\institution{Research School of Computer Science, 
 Australian National University}
\city{Canberra}
 \state{ACT}
\postcode{0200}
}
  \email{{ghodai.abdelrahman, qing.wang}@anu.edu.au}

\renewcommand{\shortauthors}{G. Abdelrahman and Q. Wang}

\begin{abstract}
 Can machines trace human knowledge like humans? Knowledge tracing (KT) is a fundamental task in a wide range of applications in education, such as massive open online courses (MOOCs), intelligent tutoring systems, educational games, and learning management systems. It models dynamics in a student's knowledge states in relation to different learning concepts through their interactions with learning activities. Recently, several attempts have been made to use deep learning models for tackling the KT problem. Although these deep learning models have shown promising results, they have limitations: either lack the ability to go deeper to trace how specific concepts in a knowledge state are mastered by a student, or fail to capture long-term dependencies in an exercise sequence. In this paper, we address these limitations by proposing a novel deep learning model for knowledge tracing, namely Sequential Key-Value Memory Networks (SKVMN). This model unifies the strengths of recurrent modelling capacity and memory capacity of the existing deep learning KT models for modelling student learning. We have extensively evaluated our proposed model on five benchmark datasets. The experimental results show that (1) SKVMN outperforms the state-of-the-art KT models on all datasets, (2) SKVMN can better discover the correlation between latent concepts and questions,  and (3) SKVMN can trace  the knowledge state of students dynamics, and a leverage sequential dependencies in an exercise sequence for improved predication accuracy.
\end{abstract}

%
%


\maketitle

\input{section_Introduction.tex}

\input{section_Model.tex}

\input{section_Experiments.tex}

\input{section_RelatedWork.tex}

\input{section_Conclusion.tex}

\bibliographystyle{ACM-Reference-Format}
\balance
\bibliography{bibliography}

\end{document}

%% file: section_Introduction.tex
\section{Introduction}\label{sec:intro}
One of the prominent features of human intelligence is the ability to track their current knowledge states in mastering specific skills or concepts. This enables humans to identify gaps in their knowledge states to personalise their learning experience. With the success of artificial intelligence (AI) in modeling various areas of human cognition \cite{lecun2015deep,NLP1,NLP2,Video1,video2}, the question has arisen: can machines trace human knowledge like humans? This motivated the study of \emph{knowledge tracing} (KT), which aims to model the knowledge states of students in mastering skills and concepts, through a sequence of learning activities they participate in \cite{Corbett1994,DKT2015_5654,DKVMN17}.

Knowledge tracing is of fundamental importance to a wide range of applications in education, such as massive open online courses (MOOCs), intelligent tutoring systems, educational games, and learning management systems. Improvements in knowledge tracing can drive novel techniques to advance human learning. For example, knowledge tracing can be used to discover students' individual learning needs so that personalised learning and support can be provided to fulfill diverse capabilities of each student \cite{khajah2016deep}. It can also be used by human experts to design new measures of student learning and new teaching materials based on learning strengths and weaknesses of students \cite{piech2016uncovering}. Nonetheless, tracing human knowledge using machines is a rather challenging task. This is due to the complexity of human learning behaviors (e.g., memorising, guessing, forgetting, etc.) and the inherent difficulties of modeling human knowledge (i.e. skills and prior background) \cite{DKT2015_5654}.

\begin{figure*}[t!]
\includegraphics[scale=0.465]{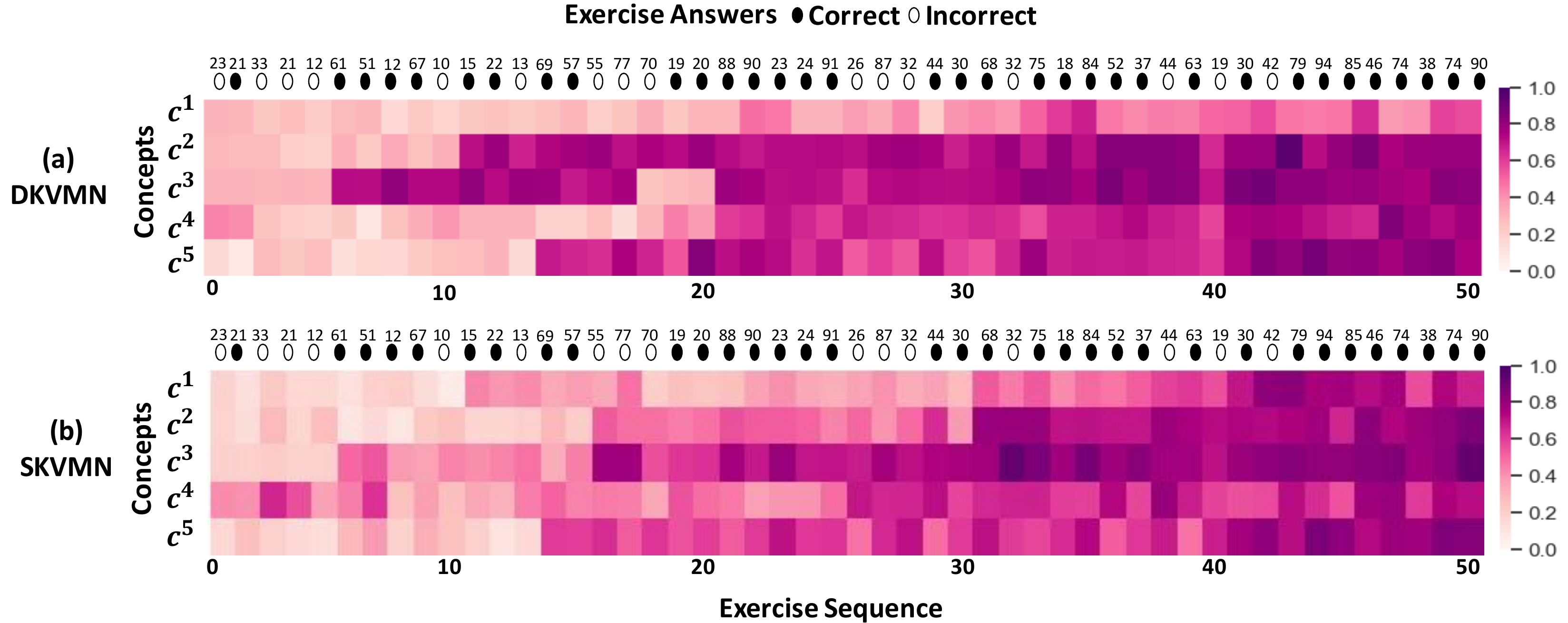}\vspace*{-0.1cm}
\caption{An illustration of how a student's knowledge states are evolving for a sequence of 50 exercises in the dataset ASSISTments2009 using: (a) DKVMN and (b) SKVMN. There are 5 concepts $\{c^1, \dots, c^5\}$ underlying these 50 exercises. Each row depicts the evolution of the concept states for a specific concept, and each column depicts the knowledge state at a certain time step. }
\label{fig:heatmap}\vspace*{-0.2cm}
\end{figure*}

Existing knowledge tracing models can be generally classified into two categories: traditional machine learning KT models and deep learning KT models. Among traditional machine learning KT models, \emph{Bayesian Knowledge Tracing} (BKT) is the most popular \citep{Corbett1994}, which models the knowledge tracing problem as predicting the state of a dynamical system that has hidden latent variables (i.e. learning concepts). In addition to BKT, probabilistic graphical models such as Hidden Markov Models (HMMs) \citep{Corbett1994,Baker2008} or Bayesian belief networks \citep{Villano92} have also been used to model knowledge tracing. To keep the inference computation tractable, traditional machine learning KT models use discrete random state variables with simple transition regimes, which limits their ability to represent complex dynamics between learning concepts. Moreover, these models often assume a first-order Markov chain for an exercise sequence (i.e. considering the most recent observation to be representing the whole history) which also limits their ability to model long-term dependencies in an exercise sequence. 

Inspired by recent advances in deep learning \citep{lecun2015deep}, several deep learning KT models have been proposed. A pioneer work by Piech et al. \citep{DKT2015_5654} reported \emph{Deep Knowledge Tracing} (DKT), which uses a Recurrent Neural Networks (RNN) with Long Short-Term Memory (LSTM) units \citep{Graves13,Sutskever_2014} to predict student performance on new exercises given their past learning history. In DKT, a student's knowledge states are represented by a sequence of hidden states that successively encode relevant information from past observations over time. Although DKT has achieved substantial improvements in prediction performance over BKT,  due to the limitation of representing a knowledge state by one hidden state, it lacks the ability to go deeper to trace how specific concepts are mastered by a student (i.e., concept states) in a knowledge state. To deal with this limitation, \emph{Dynamic Key-Value Memory Networks} (DKVMN) \citep{DKVMN17} was proposed to model a student's knowledge state as a complex function over all underlying concept states using a key-value memory. Their idea of augmenting DKVMN with an auxiliary memory follows the concepts of Memory-Augmented Neural Networks (MANN) \citep{pmlr-v48-santoro16,Graves2016}. 
However, DKVMN acquires the knowledge growth through the most recent exercise and thus fails to capture long-term dependencies in an exercise sequence (i.e., relevant past experience to a new observation). 

In this paper, we present a new KT model, called \emph{Sequential Key-Value Memory Networks} (SKVMN). This model provides three advantages over the existing deep learning KT models: 
\begin{itemize}
\item First, SKVMN unifies the strengths of both recurrent modelling capacity of DKT and memory capacity of DKVMN for modelling student learning. We observe that, although a key-value memory can help trace concept states of a student, it is not effective in modeling long-term dependencies on sequential data. We remedy this issue by incorporating LSTMs into the sequence modelling for a student's knowledge states over time. Thus, SKVMN is not only augmented with a key-value memory to enhance representation capability of knowledge states at each time step, but also can provide recurrent modelling capability for capturing dependencies among knowledge states at different time steps in a sequence. 

\item Second, SKVMN uses a modified LSTM with hops, called \emph{Hop-LSTM}, in its sequence modelling. Hop-LSTM deviates from the standard LSTM architecture by using a triangular layer for discovering sequential dependencies between exercises in a sequence. Then, the model may hop across LSTM cells according to the relevancy of the latent learning concepts. This enables relevant exercises that correlate to similar concepts to be processed together. In doing so, the inference becomes faster and the capacity of capturing long-term dependencies in an exercise sequence is enhanced.

\item Third, SKVMN improves the write process of DKVMN in order to better represent knowledge states stored in a key-value memory. In DKVMN, the current knowledge state is not considered when calculating the knowledge growth of a new exercise. This means that the previous learning experience is ignored. For example, when a student attempts the same question multiple times, the same knowledge growth would be added to the knowledge state, regardless of whether the student has previously answered this question or how many times the answers were correct. SKVMN solves this issue by using a summary vector as input for the write process, which reflects both the current knowledge state of a student and the prior difficulty of a new question. 

\end{itemize}

We have extensively evaluated our proposed model SKVMN on five well-established KT benchmark datasets, and compared it with the state-of-the-art KT models. The experimental results show that (1) SKVMN outperforms the existing KT models, including DKT and DKVMN, on all five datasets, (2) SKVMN can better discover the correlation between latent concepts and questions, and (3) SKVMN can  the knowledge state of students dynamics, and leverage sequential dependencies between exercises in an exercise sequence for improved predication accuracy.

\begin{figure*}[t!]
\hspace*{-0.3cm}\begin{minipage}{0.86\linewidth}
\hspace*{0.3cm}\includegraphics[scale=0.85]{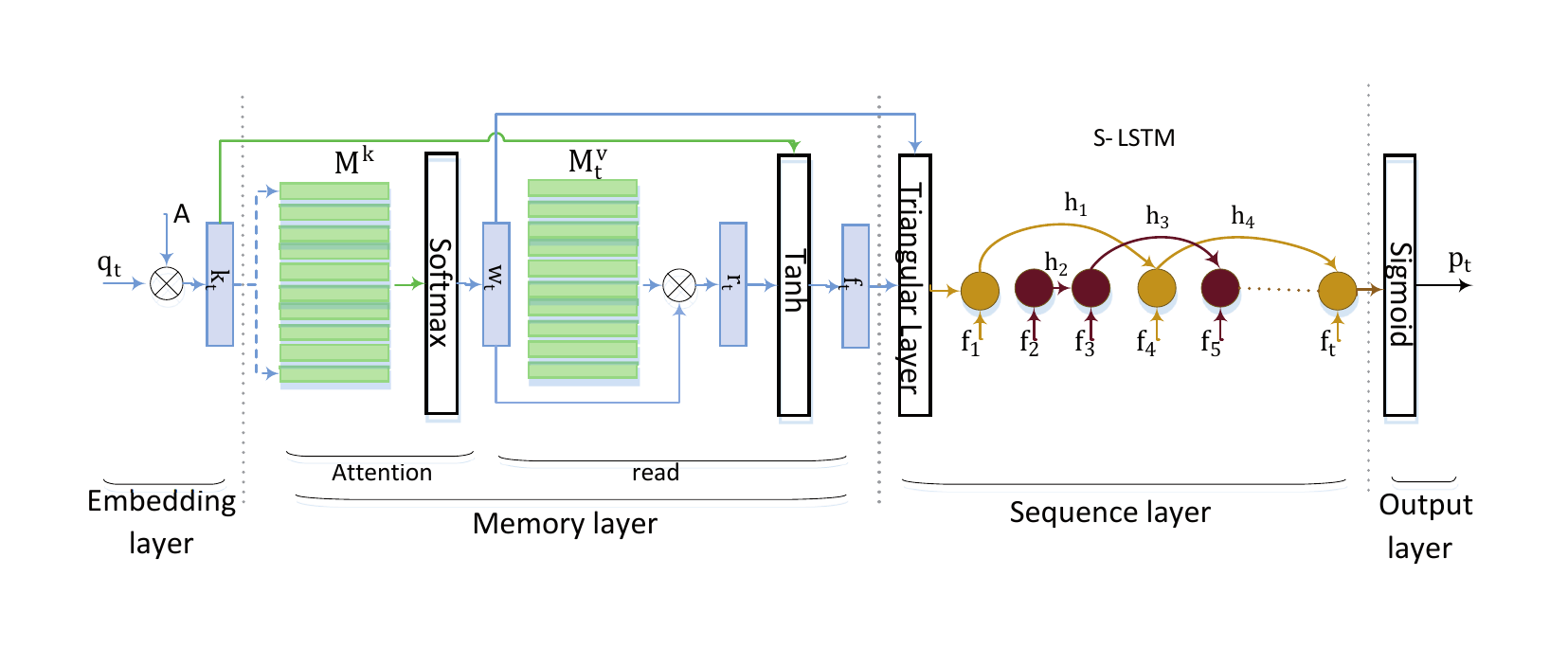}
\vspace{-0.6cm}
\begin{center}
(a)  
\end{center}
\end{minipage}
\begin{minipage}{0.13\linewidth}
\vspace*{0.3cm}
\centering
\begin{tabular}{|c|} 
\multicolumn{1}{c}{Sequential}\\
\multicolumn{1}{c}{dependencies}\\
\hline 
 $q_1\leftarrow q_4$ \\
 $q_2\leftarrow q_3$ \\
 $q_5\leftarrow q_3$ \\
   ...\\
 $q_4\leftarrow q_t$ \\
\hline 
\end{tabular}
\vspace{0.9cm}
\begin{center}
(b)  
\end{center}
\end{minipage}
\vspace{-0cm}
\caption{An illustration of the SKVMN model at time step $\mathrm{t}$, where $\mathrm{q_{t}}$ is the input question, the exercise history $\mathrm{X}=\langle \mathrm{ \left(q_{1},y_{1}\right),\left(q_{2},y_{2}\right),\dots,\left(q_{t-1},y_{t-1}\right)\rangle}$, and each $\mathbf{f_i}$ is the summary vector of the question $\mathrm{q_{i}}$ for $\mathrm{i\in [1,t]}$: (a) the model has four layers, namely the embedding, memory, sequence and output layers; and (b) sequential dependencies among questions in $\mathrm{X}$.}
\label{fig:Model_Arc}
\end{figure*}

Figure \ref{fig:heatmap} illustrates how a student's knowledge states are evolving as the student attempts a sequence of 50 exercises in DKVMN and SKVMN. We can see that, compared with the knowledge states of DKVMN depicted in Figure \ref{fig:heatmap}.(a), our model SKVMN provides a smoother transition between two successive concept states as depicted in Figure \ref{fig:heatmap}.(b). Moreover, SKVMN captures a smooth, progressive evolution of knowledge states over time (i.e., through this exercise sequence), which more accurately reflects the way a student learns (will be discussed in detail in Section \ref{subsec:evolution-knowledge-state}).

The reminder of this paper is organized as follows. Section \ref{sec:pd} defines the knowledge tracing problem. Section \ref{sec:method} presents our proposed KT model SKVMN. Sections \ref{sec:exp} and \ref{sec:res} discuss the experimental design and results. The related work is presented in Section \ref{sec:rw}. We conclude the paper in Section \ref{sec:conc}.

%% file: section_Model.tex
 \section{Problem Formulation}
\label{sec:pd}
Broadly speaking, knowledge tracing is to track down students' knowledge states over time through a sequence of observations on how the students interact with given learning activities. In this paper, we formulate knowledge tracing as a sequence prediction problem in machine learning, which learns the knowledge state of a student based on an exercise answering history.

Let $\mathrm{Q}=\left\{ q_{1},\ldots,q_{|Q|}\right\}$ be the set of all distinct question tags in a dataset.  Each  $\mathrm{q_{i}\in Q}$  may have a different level of difficulty, which is not explicitly provided. An \emph{exercise} $\mathrm{x_{i}}$ is a pair $\mathrm{\left(q_{i},y_{i}\right)}$ consisting of a question tag $\mathrm{q_{i}}$ and a binary variable $\mathrm{\mathrm{y_{i}\in\{0,1\}}}$ representing the answer, where $0$ means that $\mathrm{q_{i}}$ is incorrectly answered and $1$ means that $\mathrm{q_{i}}$ is correctly answered. When a student interacts with the questions in $\mathrm{Q}$, a history of exercises $\mathrm{X}=\langle \mathrm x_{1},x_{2},\ldots,x_{t-1}\rangle $ undertaken by the student can be observed. Based on a history of exercises $\mathrm{X}=\langle \mathrm x_{1},x_{2},\ldots,x_{t-1}\rangle $,  we want to predict the probability of correctly answering a new question at time step $\mathrm{t}$ by the student, i.e., $\mathrm{p_{t}=\left(y_{t}=1|q_{t},\mathrm{X}\right)}$. 

We assume that the questions in $\mathrm{Q}$ are associated with $\mathrm{N}$ latent concepts $\mathrm{C}$. The \emph{concept state} of each latent concept $\mathrm{c^{i}\in C}$ is a random variable describing the mastery level of the student on this latent concept. At each time step $\mathrm{t}$, the \emph{knowledge state} of a student is modelled as a set of all concept states of the student, each corresponding to a latent concept in $\mathrm{C}$ at time step $\mathrm{t}$.

\section{Sequential Key-Value Memory Networks}
\label{sec:method}
In this section, we introduce our model \emph{Sequential Key-Value Memory Networks} (SKVMN). We first present an overview for SKVMN. Then, we show how a key-value memory can be attended, read and written in our model.  To leverage sequential dependencies among latent concepts for predication, we then present a modified LSTMs, called \emph{Hop-LSTMs}. Lastly, we discuss the optimisation techniques used in the model.

\vspace{-0cm}
\subsection{Model Overview}

The SKVMN model is augmented with a key-value memory $\langle \mathbf{M^{k}}, \mathbf{M^{v}}\rangle$ following the work in \citep{DKVMN17}, i.e., a pair of one static matrix $\mathbf{M^{k}}$ of size $\mathrm{N\times d_{k}}$, called the \emph{key matrix}, and one dynamic matrix $\mathbf{M^{v}}$ of size $\mathrm{N\times d_{v}}$, called the \emph{value matrix}. Both the key matrix and the value matrix have the same $N$ memory slots, but they may differ in their state dimensions $\mathrm{d_{k}}$ and $\mathrm{d_{v}}$. The key matrix stores the latent concepts underlying questions, and the value matrix stores the concept states of a student (i.e., the knowledge state) which can be changed dynamically based on student learning.

Given an input question $\mathrm{q_{t}}$ at time step $\mathrm{t}$, the SKVMN model retrieves the knowledge state of a student from the key-value memory $\langle \mathbf{M^{k}}, \mathbf{M^{v}}\rangle$,  and predicts the probability of correctly answering the question $\mathrm{q_{t}}$ by the student.
 Figure \ref{fig:Model_Arc}.(a) illustrates the SKVMN model at time step $\mathrm{t}$, which consists of four layers: the embedding, memory, sequence and output layers.

\begin{itemize}
\item The embedding layer is responsible for mapping an input question at time step $\mathrm{t}$ into a high-dimensional vector space. 
\item The memory layer involves two processes: attention and read, where the attention process provides an addressing
mechanism for the input question $\mathrm{q_{t}}$ to allocate the relevant information from the key-value memory, and the read process uses the attention vector to retrieve the current knowledge state of the student from the value matrix $\mathbf{M^{v}_t}$. The details of the attention and read processes will be
discussed in Sections \ref{sec:attention} and \ref{sec:read}.
\item The sequence layer consists of a set of recurrently connected
LSTM cells, where LSTM cells are connected based on their sequential dependencies determined by a Triangular layer as depicted in Figure \ref{fig:Model_Arc}.(b). The details of the sequence layer will be
discussed in Section \ref{ssc:sequence-layer}.
\item The output layer generates the probability of correctly answering the input question $\mathrm{q_{t}}$.
\end{itemize}

After a student has attempted the input question $\mathrm{q_{t}}$ with the answer $\mathrm{y_{t}}$, the value matrix in the key-value memory of the SKVMN model needs to be updated in order to reflect the latest knowledge state of the student. Figure \ref{fig:write} depicts how the value matrix is transited from $\mathbf{M_{t}^{v}}$ at time step $\mathrm{t}$ to $\mathbf{M_{t+1}^{v}}$ at time step $\mathrm{t+1}$ using the write process. The details of the write process will be
discussed in Section \ref{sec:write}.

\subsection{Attention, Read and Write}
There are three processes relating to access to a key-value memory in our model: attention, read and write. In the following, we elaborate these processes.

\subsubsection{\textbf{Attention}}\label{sec:attention}
Given a question $\mathrm{q_{t}}\in Q$ as input, the model represents $\mathrm{q_{t}}$ as a ``one-hot" vector of length $\mathrm{|Q|}$ in which all entries are zero, except for the entry that corresponds to $\mathrm{q_{t}}$. 

In order to map $\mathrm{q_{t}}$ into a continuous vector space, $\mathrm{q_{t}}$ is multiplied by an embedding matrix $\mathbf{A\mathrm{\in\mathbb{R\mathrm{^{|Q|\times d_{k}}}}}}$, which generates a continuous embedding vector $\mathbf{k_{t}}\in \mathbb{R\mathrm{^{d_{k}}}}$, where $\mathrm{d_{k}}$ is the embedding dimension. Then, an attention vector $\mathbf{w_{t}}$ is obtained by applying the Softmax function to the inner product between the embedding vector $\mathbf{k_{t}}$ and each key slot $\mathbf{M^{k}\mathrm{(i)}}$ in the key matrix $\mathbf{M^{k}}$ as follows:
 \begin{equation}
\label{eq:weightVec}
\mathrm{\mathbf{w_{t}}(i)=Softmax(\mathrm{\mathbf{k_{t}^{T}}}\mathbf{M^{k}}(i))}
\end{equation}
where $\mathrm{Softmax(z_{i})=e^{z_{i}}/\sum_{j}e^{z_{j}}}$. Conceptually, $\mathbf{w_{t}}$ represents the correlation between the question $\mathrm{q_{t}}$ and the underlying latent concepts stored in the key matrix  $\mathbf{M^{k}}$.

\subsubsection{\textbf{Read}}\label{sec:read}

For each exercise $\mathrm{x_{t}=(q_{t},y_{t})}$, the model uses its corresponding attention vector $\mathbf{w_{t}}$ to retrieve the concept states of the student with regard to the question $\mathrm{q_{t}}$ from the value matrix $\mathbf{M_{t}^{v}}$. Specifically, the read process takes the attention vector $\mathbf{w_{t}}$ as input and yields a read vector $\mathbf{r_{t}}$ which is the weighted sum of all values being attended by $\mathbf{w_{t}}$ in the memory slots of the value matrix $\mathbf{M_{t}^{v}}$, i.e., 
\begin{equation}
\label{eq:readVec}
\mathbf{r_{t}}=\mathrm{\sum_{i=1}^{N}}\mathbf{\mathrm{w}_{t}}\mathrm{(i)}\mathbf{M_{t}^{v}\mathrm{(i)}}
\end{equation}

The read vector $\mathbf{r_{t}}$ is concatenated with the embedding vector $\mathbf{k_{t}}$. Then, the combined vector is fed to a Tanh layer to calculate the summary vector $\mathbf{f_{t}}$:
\begin{equation}
\label{eq:summaryVec}
\mathrm{\mathbf{f_{t}}=Tanh(\mathbf{W_{\mathrm{1}}^{\mathrm{T}}\mathrm{\left[\mathbf{r_{t},k_{t}}\right]+b_{1})}}}
\end{equation}   
where $\mathrm{Tanh(z_{i})=(e^{z_{i}}-e^{-z_{i}})/(e^{z_{i}}+e^{-z_{i}})}$, $\mathbf{W}_1$ is the weight matrix of the Tanh layer, and $\mathbf{b}_1$ is the bias vector. While the read vector $\mathbf{r_{t}}$ represents the student's knowledge state with respect to the relevant concepts of the current question $\mathrm{q_{t}}$, the summary vector $\mathbf{f_{t}}$ adds prior information of the question (e.g., the level of difficulty) to this knowledge state. Intuitively, $\mathbf{f_{t}}$ represents how well the student has mastered the latent concepts relevant to the question $\mathrm{q_{t}}$ before attempting this question. 

\begin{figure}[t!]
\includegraphics[scale=0.65]{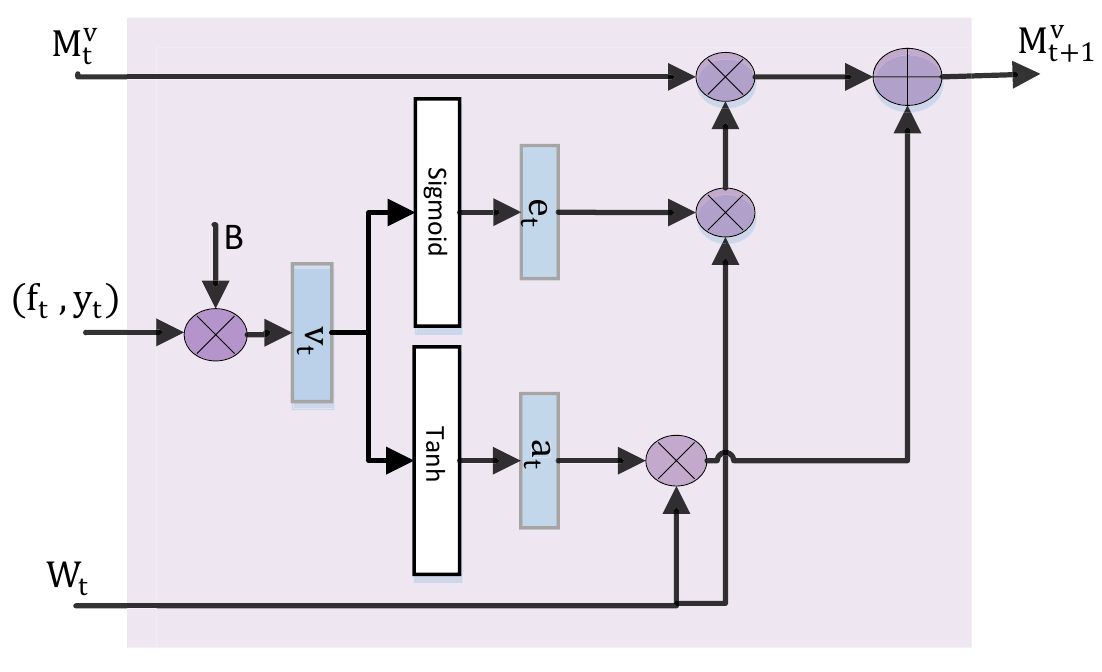}
\caption{An illustration of the write process in the SKVMN model, which transits the value matrix from $\mathbf{M_{t}^{v}}$ at time step $\mathrm{t}$ to $\mathbf{M_{t+1}^{v}}$ at time step $\mathrm{t+1}$, where $\mathrm{(\mathbf{f_{t}},y_{t})}$ and $\mathbf{w_t}$ are the input to the write process at time step $\mathrm{t}$.}
\label{fig:write}
\end{figure}

\subsubsection{\textbf{Write}}\label{sec:write}
The write process occurs each time after the student has attempted a question. The purpose of this write process is to update the concept states of the student in the value matrix $\mathbf{M_{t}^{v}}$ using the knowledge growth gained through attempting the question $\mathrm{q_{t}}$. This update leads to the transition of the value matrix from from $\mathbf{M_{t}^{v}}$ to $\mathbf{{M}_{t+1}^{v}}$ as depicted in Figure \ref{fig:write}.

To calculate the knowledge growth, our model considers not only the correctness of the answer $\mathrm{y_{t}}$ for $\mathrm{q_{t}}$, but also the student's mastery level of the concept states $\mathbf{f_{t}}$ before attempting $\mathrm{q_{t}}$. Each $\mathrm{(\mathbf{f_{t}},y_{t})}$ is represented as a vector of length $\mathrm{2|\mathrm{Q}|}$ and multiplied by an embedding matrix $\mathbf{B\mathrm{\in\mathbb{R\mathrm{^{2|Q|\times d_{v}}}}}}$ to get a write vector $\mathbf{v_{t}}$, which represents the knowledge growth of the student obtained by attempting the question. 

Similar to other memory augmented networks \citep{Graves2016, DKVMN17}, the write process proceeds with two gates: \emph{erase gate} and \emph{add gate}. The former controls what information to erase from the knowledge state after attempting the latest exercise $\mathrm{x_{t}}$, while the latter controls what information to add into the knowledge state. From the knowledge tracing perspective, these two gates capture the forgetting and enhancing aspects of learning knowledge, respectively.
 
With the write vector $\mathbf{v_{t}}$ for the knowledge growth, an erase vector $\mathbf{e_{t}}$ is calculated as:
 \begin{equation}
 \label{eq:erase_signal}
\mathrm{\mathbf{e_{t}}=sigmoid(\mathbf{W_{\mathrm{e}}^{\mathrm{T}}\cdot\mathrm{\mathbf{v_{t}}+b_{e})}}}
\end{equation}   
where $\mathrm{\mathrm{Sigmoid(z_{i})=1/(1+e^{-z_{i}})}}$ and $\mathbf{W}_e$ is the weight matrix. Using the attention vector $\mathbf{w_{t}}$, the value matrix $\mathbf{\tilde{M}_{t+1}^{v}}$ after applying the erase vector $\mathbf{e_{t}}$ is
\begin{equation}
\label{eq:Erased_Memory}
\mathbf{\tilde{M}_{t+1}^{v}\mathrm{(i)}}=\mathbf{M_{t}^{v}}\mathrm{(i)[1-\mathbf{w_{t}}(i)\mathbf{e_{t}}]}.
\end{equation}

Then, an add vector $\mathbf{a_{t}}$ is calculated by
\begin{equation}
\label{eq:Add_signal}
\mathrm{\mathbf{a_{t}}=Tanh(\mathbf{W_{\mathrm{a}}^{\mathrm{T}}\cdot\mathrm{\mathbf{v_{t}}+b_{a})}}}
\end{equation}  
where $\mathbf{W}_a$ is the weight matrix. Finally, the value matrix $\mathbf{M_{t+1}^{v}}$ for the next knowledge state is updated as:
\begin{equation}
\mathbf{M_{t+1}^{v}\mathrm{(i)}}=\mathbf{\tilde{M}_{t+1}^{v}}\mathrm{(i)+\mathbf{w_{t}}(i)\mathbf{a_{t}}}
\end{equation} 

\subsection{Sequence Modelling}\label{ssc:sequence-layer} Now we discuss the sequence modelling approach used at the sequence layer for predicting the probability of correctly answering the question $\mathrm{q_{t}}$ based on the exercise history.

\subsubsection{\textbf{Sequential dependencies}}

An exercise history $\mathrm{X}$ of a student may contain a long sequence of exercises, for example, the average sequence length in the ASSISTments2009 dataset is $233\pm100$ questions per sequence. However, as different exercises may correlate to different latent concepts, not all exercises in $\mathrm{X}$ can equally contribute to the prediction of answering a given question $\mathrm{q_{t}}$. Thus, we observe that, by hopping across irrelevant exercises in $\mathrm{X}$ with regard to $\mathrm{q_{t}}$, recurrent models can be applied on a shorter and more relevant sequence, leading to more efficient and accurate prediction performance. 

For each question in $\mathrm{Q}$, since its attention vector reflects the correlation between this question and the latent concepts in $\mathrm{C}$, we consider that the similarity between two attention vectors can provide a good indication of how their corresponding questions are relevant in terms of their correlations with latent concepts. For example, suppose that we have three latent concepts $\mathrm{C=\{c^{1},c^{2},c^{3}\}}$, and two attention vectors $\mathrm{\mathbf{w_{1}}=[0.15,0.25,0.6]^{T}}$
 and $\mathrm{\mathbf{w_{2}}=[0.2,0.3,0.5]^{T}}$ that correspond to the questions $\mathrm{q_{1}}$ and $\mathrm{q_{2}}$, respectively.  Then $\mathrm{q_{1}}$ and $\mathrm{q_{2}}$ are considered as being relevant if both $\mathbf{w_{1}}$ and $\mathbf{w_{2}}$ are mapped to a vector $\mathrm{[0,0,1]^{T}}$ where $0$, $1$ and $2$ refer to the value ranges ``low", ``middle" and ``high", respectively.

Now, the question is: how to identify similar attention vectors? For this, we use the triangular membership function \citep{klir1995fuzzy}:
\begin{equation}
\label{eq:TraingularFunc}
\mu(x)= max(min(\frac{x-a}{b-a}, \frac{c-x}{c-b}),0),
\end{equation}
where the parameters $a$ and $c$ determine the feet and the parameter $b$ determines the peak of a triangle. We use three triangular membership functions for three value ranges: low (0), medium (1), and high (2). Each real-valued component in an attention vector is mapped to one of the three value ranges. Each attention vector $\mathbf{w_t}$ is associated with an identity vector $\mathbf{d_t}$. Similar attention vectors have the same identity vector, while dissimilar attention vectors have different identity vectors.

Then, at each time step $\mathrm{t}$, for the current question $\mathrm{q_{t}}$ and an exercise history $\mathrm{X}$, we say $\mathrm{q_{t}}$ is \emph{sequentially dependent} on $\mathrm{q_{t-\lambda}}$ in $\mathrm{X}$ with $\lambda\in (0,t)$, denoted as $\mathrm{q_{t-\lambda}}\leftarrow \mathrm{q_{t}}$, if the following two conditions are satisfied: 
\begin{itemize}
\item The attention vectors of $\mathrm{q_{t-\lambda}}$ and $\mathrm{q_{t}}$ have the same identity vector, i.e., $\mathbf{d}_{t-\lambda}=\mathbf{d_t}$, and 
\item There is no other $\mathrm{q_{j}}$ in $\mathrm{X}$ such that $\mathbf{d_j}=\mathbf{d_t}$ and $\mathrm{j>t}-\lambda$, i.e., $\mathrm{q_{t-\lambda}}$ is the most recent exercise in $\mathrm{X}$ that is relevant to $\mathrm{q_{t}}$. 
\end{itemize}Over time, given a sequence of questions $\langle \mathrm{q_{1}},\mathrm{q_{2}},\dots, \mathrm{q_{|Q|}}\rangle$, we thus have an exercise history $\mathrm{X'}$ in which exercises are partitioned into a set of subsequences $\mathrm{\{X'_1, \dots, X'_n\}}$ with $\mathrm{\Sigma_{k=1}^n|X'_k|=|X'|}$ and, for any two consecutive exercises $\mathrm{\left(q_{i},y_{i}\right)}$ and $\mathrm{\left(q_{j},y_{j}\right)}$ in the subsequence $\mathrm{X'_{k}}$, $\mathrm{q_{j}}$ is sequentially dependent on $\mathrm{q_{i}}$, i.e. $\mathrm{q_{i}\leftarrow q_{j}}$. In Figure \ref{fig:Model_Arc}, based on the sequential dependencies presented in Figure \ref{fig:Model_Arc}.(b), $\mathrm{X}$ is partitioned into $\mathrm{\langle q_1, q_4, \dots, q_t \rangle}$ and $\mathrm{\langle q_2, q_3, q_5, \dots \rangle}$, which correspond to the recurrently connected LSTM cells depicted in Figure \ref{fig:Model_Arc}.(a). We will discuss further details in the following.

\subsubsection{\textbf{Hop-LSTM}} 

Based on sequence dependencies between exercises in a sequence, a modified LSTM with hops, called Hop-LSTM, is used to predict the probability of correctly answering a new question by a student. Different from the standard LSTM architecture, Hop-LSTM allows us to recurrently connect the LSTM cells for questions based on the relevance of their latent concepts. More precisely, two LSTM cells in Hop-LSTM are connected only if the input question of one LSTM cell is sequentially dependent on the input question of the other LSTM cell. This means that Hop-LSTM has the capability of hopping across the LSTM cells when their input questions are irrelevant to the current question $\mathrm{q_{t}}$. 

Formally, at each time step $\mathrm{t}$, for the question $\mathrm{q_{t}}$, if there is an exercise $\mathrm{\left(q_{t-\lambda},y_{t-\lambda}\right)}\in \mathrm{X}$ with $\lambda \in (0,t)$ and $\mathrm{q_{t-\lambda}\leftarrow q_t}$, then the current LSTM cell takes the summary vector $\mathbf{f}_{t}$ and the hidden state $\mathbf{h}_{t-\lambda}$ as input. Moreover, this LSTM cell updates the cell state $\mathbf{c}_{t-\lambda}$ into the cell state $\mathbf{c}_{t}$, and generates the new hidden state  $\mathbf{h}_{t}$ as output. As in \citep{lstm97}, the LSTM cell used in our work has three gates: \emph{forget gate} $\mathbf{g}_{t}$, \emph{input gate} $\mathbf{i}_{t}$, and \emph{output gate} $\mathbf{o}_{t}$ in addition to the hidden state $\mathbf{h}_{t}$ and the cell state $\mathbf{c}_{t}$:
\begin{align} 
&\mathbf{g}_t=Sigmoid(\mathbf{W}_g[\mathbf{h}_{t-\lambda},\mathbf{f}_t]+\mathbf{b}_g) \\ 
&\mathbf{i}_t=Sigmoid(\mathbf{W}_i[\mathbf{h}_{t-\lambda},\mathbf{f}_t]+\mathbf{b}_i)\\
&\mathbf{o}_t=Sigmoid(\mathbf{W}_o[\mathbf{h}_{t-\lambda},\mathbf{f}_t]+\mathbf{b}_o)\\
&\mathbf{\tilde{c}}_t=Tanh(\mathbf{W}_c[\mathbf{h}_{t-\lambda},\mathbf{f}_t]+\mathbf{b}_c)\\
&\mathbf{c}_t=\mathbf{g}_t\odot \mathbf{c}_{t-\lambda}+\mathbf{i}_{t}\odot\mathbf{\tilde{c}}_{t}\\
&\mathbf{h}_t=\mathbf{o}_t\odot Tanh(\mathbf{c}_t)
\end{align}

Then, the output vector $\mathbf{h}_{t}$ of the curent LSTM cell is sent to a Sigmoid layer, which calculates the probability $\mathrm{p_{t}}$ of correctly answering the current question $\mathrm{q_{t}}$ by
\begin{align}
\label{eq:Final_Prob}
&p_t=Sigmoid(\mathbf{W}_2^{T}\cdot\mathbf{h}_t+\mathbf{b}_2).
\end{align}

\subsection{Model Optimisation}

To optimise the model, we use the cross-entropy loss function between the predicted probability of being correctly answered $\mathrm{p_{t}}$ and the true answer $\mathrm{y_{t}}$. The following objective function is defined over training data:
\begin{equation}
\label{eq:loss}
\mathcal{L\mathrm{\mathrm{=-\sum_{t}(y_{t}\log p_{t}+(1-y_{t})\log(1-p_{t}))}}}
\end{equation}

We initialise of the memory matrices ($\mathbf{M_{t}^{k}}$ and $\mathbf{M_{t}^{v}}$) and embedding matrices ($\mathbf{A}$ and $\mathbf{B}$) using a random Gaussian distribution $N(0,\sigma)$. While for weights and biases of the neural layers, we use Glorot uniform random initialization \citep{glorot2010} for a faster convergence. These randomly initialized parameters are optimized using the stochastic gradient decent (SGD) mechanism \citep{Bottou2012}. As we use Hop-LSTM at the sequence layer, during the backpropagation of the gradients, only the parameters of the connected LSTM cells (i.e. the ones responsible for the current prediction error) are updated. Other parameters, such as the embedding matrices, weight matrices, and bias vectors, are updated in each backpropagation iteration based on loss function values.

 Note that, through training, the model can discover relevant latent concepts for each question and store their state values in the value matrix $\mathbf{M^{v}}$.

%% file: section_Experiments.tex
\section{Experiments}
\label{sec:exp}
In this section, we present the experiments of evaluating our proposed model SKVMN against the state-of-the-art KT models. These experiments aim to answer the following research questions:
\begin{itemize}
\item[RQ1:] What is the optimal size for a key-value memory (i.e., the key and value matrices) of SKVMN?
\item[RQ2:] How does SKVMN perform on predicting a student's answers of new questions, given an exercise history? 
\item[RQ3:] How does SKVMN perform on discovering the correlation between latent concepts and questions?
\item[RQ4:] How does SKVMN perform on capturing the evolution of a student's knowledge states?
\end{itemize}
       
\subsection{Datasets}
We use five  well-established datasets in the KT literature \citep{khajah2016deep,DKVMN17,DKT2015_5654}. Table \ref{tbl:datasets} summarizes the statistics of the data sets.    

\begin{itemize}
\item \textbf{Synthetic-5\footnote{Synthetic-5:https://github.com/chrispiech/DeepKnowledgeTracing/tree/master
/data/synthetic}:} This dataset consists of two subsets: one for training and one for testing. Each subset contains $50$ distinct questions which were answered by $4,000$ virtual students. A total number of $200,000$ exercises (i.e. $\mathrm{(q_{t},y_{t})}$) are contained in the dataset.
\item \textbf{ASSISTments2009\footnote{ASSISTments2009:https://sites.google.com/site/assistmentsdata/home/assistment-2009-2010-data/skill-builder-data-2009-2010}:} This dataset was collected during the school year $2009-2010$ using the ASSISTments online education website \footnote{https://www.assistments.org/}. The dataset consists of $110$ distinct questions answered by $4,151$ students which gives a total number of $325,637$ exercises.
\item \textbf{ASSISTments2015\footnote{ASSISTments2015:https://sites.google.com/site/assistmentsdata/home/2015-assistments-skill-builder-data}:} As an update to the ASSISTments2009 dataset, this dataset was released in $2015$. It includes $100$ distinct questions answered by $19,840$ students with a total number of $683,801$ exercises. This dataset has the largest number of students among the other datasets. Albeit, the average number of exercises per student is low. The original dataset also has some incorrect answer values (i.e. $\mathrm{\mathrm{y_{i}\notin\{0,1\}}}$), which are removed during preprocessing.
\item \textbf{Statics2011\footnote{Statics2011:https://pslcdatashop.web.cmu.edu/
DatasetInfo?datasetId=507}:} This datasets was collected from a statistics course at Carnegie Mellon University during Fall $2011$. It contains $1,223$ distinct questions answered by $333$ undergraduate students with a total number of $189,297$ exercises. This dataset has the highest exercise per student ratio among all datasets.
\item \textbf{JunyiAcademy\footnote{Junyi2015: https://datashop.web.cmu.edu/DatasetInfo?datasetId=1198}:} This dataset was collected from Junyi Academy \footnote{https://www.junyiacademy.org/}, which is an education website of providing learning materials and exercises on various scientific courses, on 2015 \citep{chang2015modeling}. It contains $722$ distinct questions answered by $199,549$ students with a total number of $25,628,935$ exercises. It is the largest dataset in terms of the number of exercises.

\end{itemize}

 \begin{table}
  \caption{Dataset statistics }
  \resizebox{\columnwidth}{!}{%
  \label{tbl:datasets}
  \begin{tabular}{l|ccc|c}
    \toprule
    \multirow{2}{*}{Dataset}&\multirow{2}{*}{\#Questions} &\multirow{2}{*}{\#Students}&\multirow{2}{*}{\#Exercises}&\#Exercises\\
    & & & & per student\\
    \midrule
    Synthetic-5&$50$&$4,000$&$200,000$&$50$\\
    ASSISTments2009&$110$&$4,151$&$325,637$&$78$\\
    ASSISTments2015&$100$&$19,840$&$683,801$&$34$\\
    Statics2011&$1,223$&$333$&$189,297$&$568$\\
    JunyiAcademy&$722$&$199,549$&$25,628,935$&$128$\\
  \bottomrule
\end{tabular}
}\vspace{0.4cm}
\caption{Comparison of SKVMN with DKVMN under different numbers of memory slots $\mathrm{N}$ and state dimensions $\mathrm{d}$, where $\mathrm{m}$ refers to the number of parameters in each setting. 
}
\resizebox{\columnwidth}{!}{%
\begin{tabular}{c|cc|cc|cc}
\toprule 
\multirow{2}{*}{Dataset} & \multirow{2}{*}{$\mathrm{d}$} & \multirow{2}{*}{$\mathrm{N}$} & \multicolumn{2}{c|}{SKVMN} & \multicolumn{2}{c}{DKVMN}\tabularnewline
\cmidrule{4-7} \cmidrule{5-7} \cmidrule{6-7} \cmidrule{7-7} 
 &  &  & AUC (\%) & m &  AUC (\%) & m\tabularnewline
\midrule
\multirow{4}{*}{Synthetic-5} & 10 & 50 & 83.11 & 15K & 82.00 & 12k\tabularnewline
 & 50 & 50 & 83.67 & 30k & 82.66 & 25k\tabularnewline
 & 100 & 50 & \textbf{84.00} & 57k & \textbf{82.73} & 50k\tabularnewline
 & 200 & 50 & 83.73 & 140k & 82.71 & 130k\tabularnewline
\midrule
\multirow{4}{*}{ASSISTments2009} & 10 & 10 & \textbf{83.63} & 7.8k & 81.47 & 7k\tabularnewline
 & 50 & 20 & 82.87 & 35k & \textbf{81.57} & 31k\tabularnewline
 & 100 & 10 & 82.72 & 71k & 81.42 & 68k\tabularnewline
 & 200 & 20 & 82.63 & 181k & 81.37 & 177k\tabularnewline
\midrule
\multirow{4}{*}{ASSISTments2015} & 10 & 20 & \textbf{74.84} & 16k & \textbf{72.68} & 14k\tabularnewline
 & 50 & 10 & 74.50 & 31k & 72.66 & 29k\tabularnewline
 & 100 & 50 & 74.24 & 66k & 72.64 & 63k\tabularnewline
 & 200 & 50 & 74.20 & 163k & 72.53 & 153k\tabularnewline
\midrule
\multirow{4}{*}{Statics2011} & 10 & 10 & 84.50 & 92.8k & 82.72 & 92k\tabularnewline
 & 50 & 10 & \textbf{84.85} & 199k & \textbf{82.84} & 197k\tabularnewline
 & 100 & 10 & 84.70 & 342k & 82.71 & 338k\tabularnewline
 & 200 & 10 & 84.76 & 653k & 82.70 & 649k\tabularnewline
\midrule
\multirow{4}{*}{JunyiAcademy } & 10 & 20 & 82.50 & 16k & 79.63 & 14k\tabularnewline
 & 50 & 10 & 82.41 & 31k & 79.48 & 29k\tabularnewline
 & 100 & 50 & \textbf{82.67} & 66k & 79.54 & 63k\tabularnewline
 & 200 & 50 & 82.32 & 163k & \textbf{80.27} & 153k\tabularnewline
\bottomrule
\end{tabular}
}
\label{tbl:empirical}
\end{table}

\subsection{Baselines}
In order to evaluate the performance of our proposed model, we select the following three KT models as the baselines: 
\begin{itemize}
\item[--] Bayesian knowledge tracing (BKT) \citep{Corbett1994}, which is based on Bayesian inference in which a knowledge state is modelled as a set of binary variables, each representing the understanding of a single concept. 
\item[--] Deep knowledge tracing (DKT) \citep{DKT2015_5654} which uses recurrent neural networks to model student learning. 
\item[--] Dynamic key-value memory networks (DKVMN)  \citep{DKVMN17} which extends the memory-augmented neural networks (MANN) by a key-value memory and is considered as the state-of-the-art model for knowledge tracing. 
\end{itemize}Our proposed model is referred to as SKVMN in the experiments.

\subsection{Measures}
We use the area under the Receiver Operating Characteristic (ROC) curve, referred to as AUC \citep{ling2003auc}, to measure the prediction performance of the KT models. The AUC ranges in value from 0 to 1. An AUC score of $0.5$ means random prediction (i.e. coin flipping). The higher an AUC score goes above $0.5$, the more accurately a predictive model can perform.

\begin{figure*}[t!]
\includegraphics[height=3cm, width=18cm]{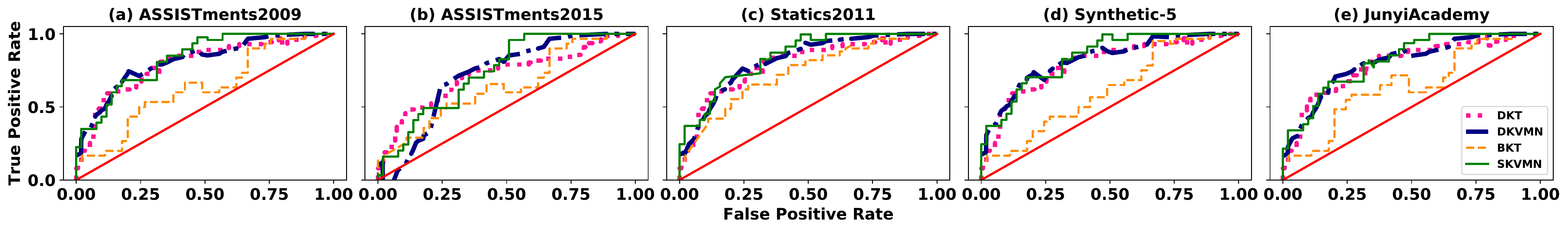}
\caption{The ROC curve results of the four models BKT, DKT, DKVMN, and SKVMN over five datasets: (a) ASSISTments2009, (b) ASSISTments2015, (c) Statics2011, (d) Synthetic-5, and (e) JunyiAcademy.}
\label{fig:AUC_res}
\end{figure*}

\subsection{Evaluation Settings}
We divided each dataset into $70$\% for training and validation and $30$\% for testing, except for Synthetic-5. This is because, as mentioned before, Synthetic-5 itself contains the training and test subsets of the same size. For each training and validation subset, we further divided it using the 5-fold cross validation (e.g. $80$\% for training and $20$\% for validation). The validation subset was used to determine the optimal values for the hyperparameters, including the memory slot dimensions $\mathrm{\mathrm{d}_{k}}$ for the key matrix and $\mathrm{\mathrm{d}_{v}}$ for the value matrix. 

\begin{table}[hbt!]
  \caption{The AUC results of the four models BKT, DKT, DKVMN, and SKVMN over all datasets.} 
  \resizebox{\columnwidth}{!}{%
  \begin{tabular}{lcccl}
    \toprule
    Dataset&BKT&DKT&DKVMN&SKVMN\\
    \midrule
    Synthetic-5&$62.0\pm0.02$&$80.3\pm0.1$&$82.7\pm0.1$&$\mathbf{84.0\pm0.04}$\\
    ASSISTments2009&$63.1\pm0.01$&$80.5\pm0.2$&$81.6\pm0.1$&$\mathbf{83.6\pm0.06}$\\
    ASSISTments2015&$64.2\pm0.03$&$72.5\pm0.1$&$72.7\pm0.1$&$\mathbf{74.8\pm0.07}$\\
    Statics2011&$73.0\pm0.01$&$80.2\pm0.2$&$82.8\pm0.1$&$\mathbf{84.9\pm0.06}$\\
    JunyiAcademy&$65.0\pm0.02$&$79.2\pm0.1$&$80.3\pm0.4$&$\mathbf{82.7\pm0.01}$\\
  \hline
\end{tabular}
}
\label{tbl:AUC}
\end{table}\vspace*{-0.3cm}

We utilised the Adam optimizer \citep{Adam2015} for SGD implementation with momentum of $0.9$ and learning rate $\gamma$ of $0.01$ annealed using a cosine function every $15$ epochs for $120$ epochs, then it remains fixed at $0.001$. The LSTM gradients were clipped to improve the training \citep{pascanu2013difficulty}. For the other baselines, we follow the optimisation procedures indicated in the original work for each of them \citep{Corbett1994,DKVMN17,DKT2015_5654}. 

A mini-batch of $32$ is selected during the training for all datasets, except Synthetic-5, for which we use a mini-batch of $8$ due to the relatively small number of training samples (i.e. exercises) in the dataset \citep{Bengio2012}. For each dataset, the training process is repeated five times, each time using a different initialization. We report the average test AUC and the standard deviation over these five runs.   

For the Triangular layer, the hyper-parameter values (${a,b,c}$) of each triangular membership function are set based on the empirical analysis of each dataset.

\section{Results and Discussion}
In this section, we present the experimental results and discuss our observations from the obtained results.

\label{sec:res}
\subsection{Hyperparameters $\mathrm{N}$ and $\mathrm{d}$}

To explore how the sizes of the key and value matrices can affect the model performance, we have conducted experiments to compare SKVMN with DKVMN under different numbers of memory slots $\mathrm{N}$ and state dimensions $\mathrm{d}$, where $\mathrm{d=d_{k}=d_{v}}$ so as to be consistent with the previous work \cite{DKVMN17}. In order to allow a fair comparison between the models SKVMN and DKVMN, we select the same set of state dimensions, (i.e. $\mathrm{d}=10,50,100,200$) and the same corresponding numbers of memory slots $\mathrm{N}$  on the datasets Synthetic-5, ASSISTments2009, ASSISTments2015 and Statics2011 to report the AUC results, following the settings originally reported in \cite{DKVMN17}. For the dataset JunyiAcademy, it was not considered in the previous work \cite{DKVMN17}. Considering that JunyiAcademy has the largest numbers of students and exercises among all datasets, we use the same settings for the numbers of memory slots and state dimensions as the ones for the second largest dataset ASSISTments2015. Table \ref{tbl:empirical} presents the AUC results for all five datasets.

As shown in Table \ref{tbl:empirical},  compared with DKVMN, our model SKVMN can produce better AUC results with comparable parameters on the datasets Synthetic-5, ASSISTments2015 and Statics2011, and with fewer parameters on the datasets ASSISTments2009 and JunyiAcademy. Particularly, for the dataset ASSISTments2009, SKVMN yields an AUC at 83.63\% with N=10, d=10 and m=7.8k, whereas DKVMN yields an AUC at 81.57\% with N=20, d=50 and m=31k (nearly 4 times of 7.8k). Similarly, for the dataset JunyiAcademy, SKVMN yields an AUC at 82.67\% with N=50, d=100 and m=66k, whereas DKVMN yields an AUC at 80.27\% with N=50, d=200 and m=153k (more than twice of 66k). 

Note that, the optimal value of $\mathrm{N}$ for ASSISTments2015 is higher than the one for its previous version (i.e. ASSISTments2009). This implies that the number of latent concepts $\mathrm{N}$ increases in ASSISTments2015 in comparison to ASSISTments2009. Moreover, the optimal value of $\mathrm{d}$ generally reflects the complexity of the exercises in a dataset, and the dataset JunyiAcademy has exercises of higher complexity than other real-world datasets.

\subsection{Prediction Accuracy}
We have conducted experiments on comparing the AUC results of our model SKVMN with the other three KT models: BKT, DKT, and DKVMN.
Table \ref{tbl:AUC} presents the AUC results of all the models. It can be seen that our model SKVMN outperformed the other models over all the five datasets. Particularly, the SKVMN model achieved an average AUC value that is at least 2\% higher than the state-of-art model DKVMN on all real-world datasets ASSISTments2009, ASSISTments2015, Statics2011, and JunyiAcademy. Even for the only synthetic dataset (i.e. Synthetic-5), the SKVMN model achieved an average AUC value of $84.0\pm0.04$, in comparison with $82.7\pm0.1$ achieved by DKVMN.
Note that the AUC values on ASSISTments2015 are the lowest among all datasets, regardless of the KT models. This reflects the difficulty of the KT task in this dataset due to its lowest exercise per student ratio, which not only makes the training process more difficult but also limits the effective use of sequence information to enhance the prediction performance. 
Figure \ref{fig:AUC_res} illustrates the ROC curves of these four models for each dataset.

\begin{figure}[t!]
\begin{minipage}{0.48\linewidth}
\centering
\hspace*{0cm}\fbox{\includegraphics[scale=0.393]{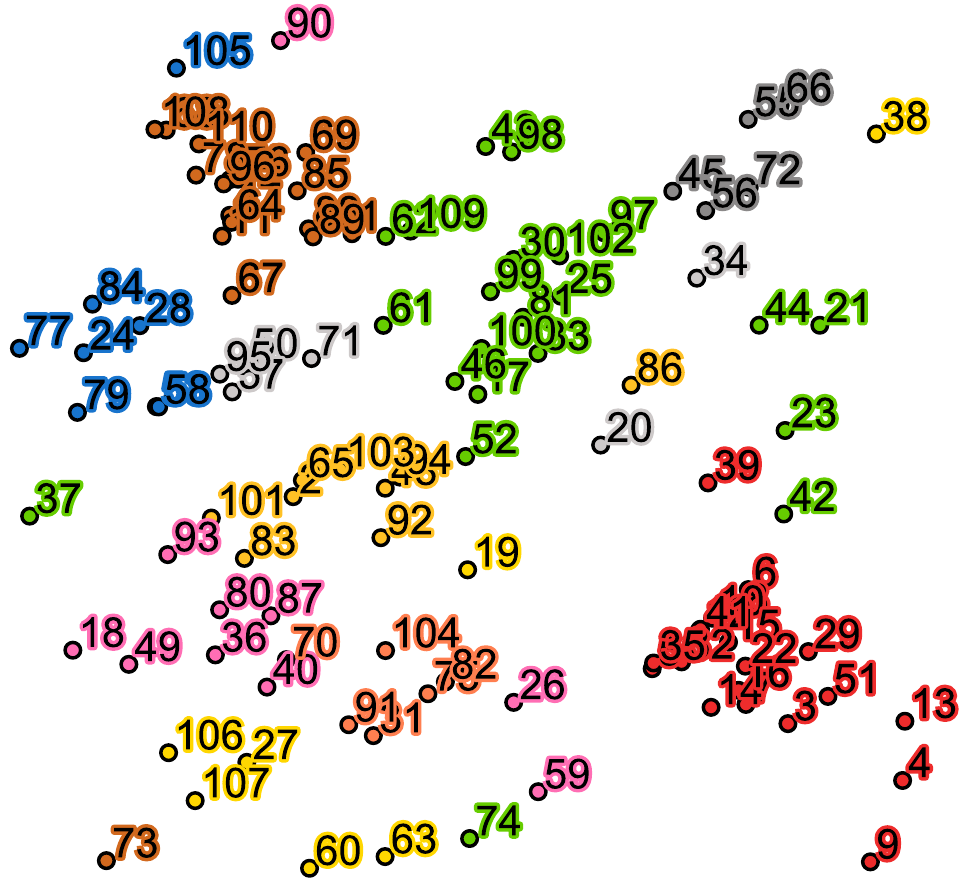}}
\vspace{-0cm}
\begin{center}
(a)  
\end{center}
\end{minipage}
\begin{minipage}{0.48\linewidth}
 \centering
\hspace*{0.1cm}\fbox{\includegraphics[scale=0.4]{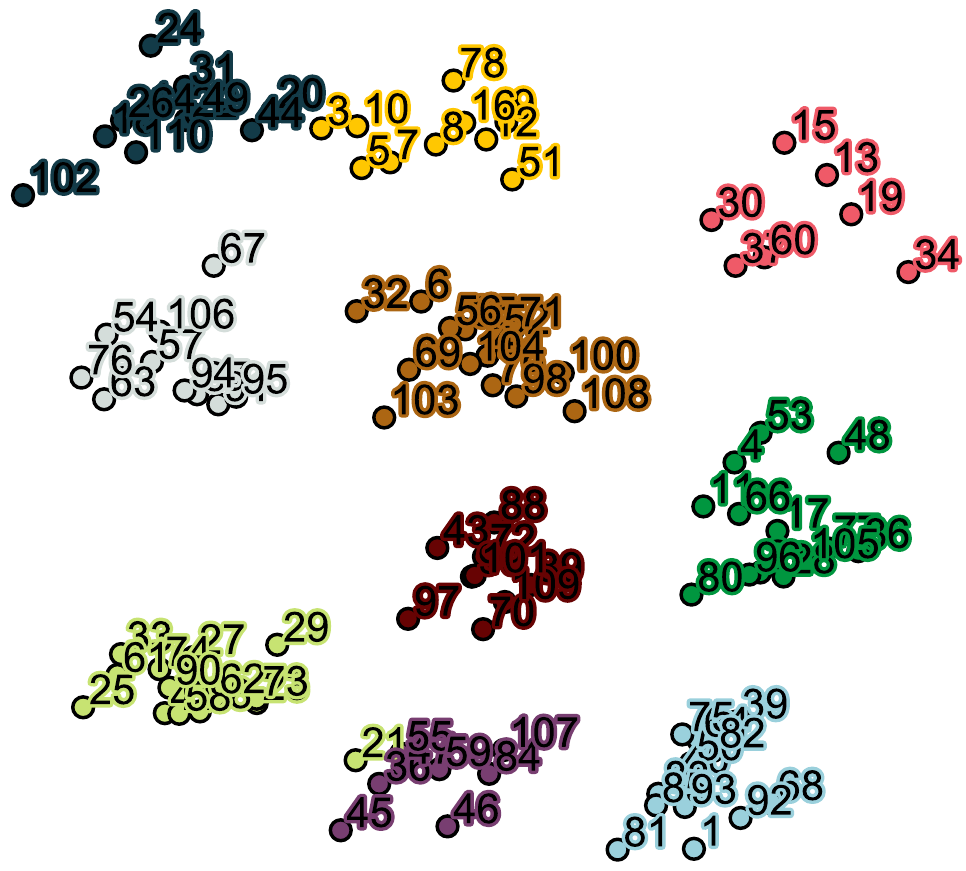}}
\vspace{-0cm}
\begin{center}
(b)  
\end{center}
\end{minipage}
\vspace{-0cm}
\caption{Clustering results of questions in the dataset ASSISTments2009 using: (a) DKVMN, and (b) SKVMN, where questions in the same color are correlating to the same latent concept.}
\label{fig:clusters}
\end{figure}
\begin{table}
\caption{Question descriptions in the dataset ASSISTments2009, where the questions are clustered according to latent concepts being discovered by SKVMN.}
\includegraphics[scale=0.40]{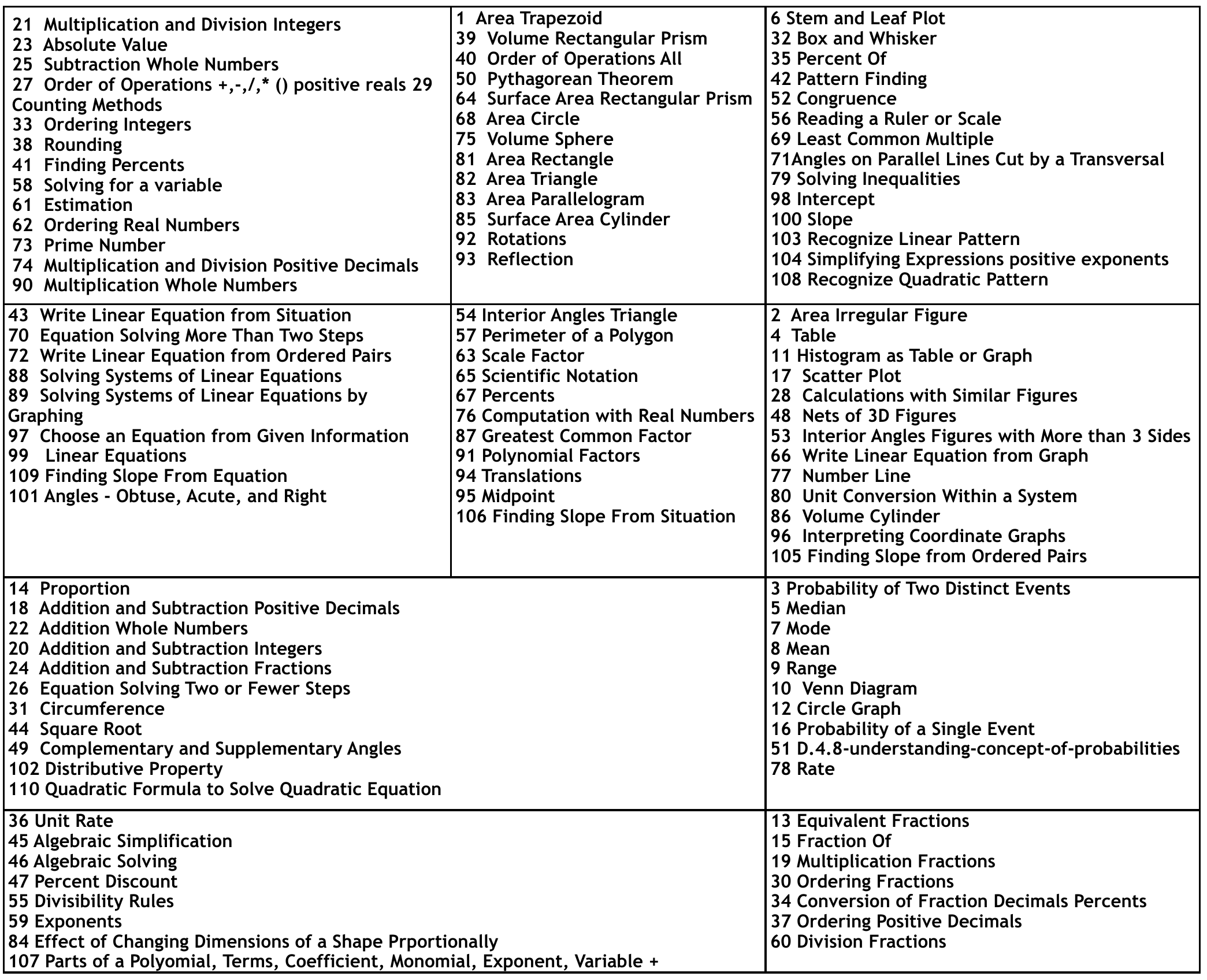}\label{tab:concepts}
\end{table}

In a nutshell, based on the AUC results in Table \ref{tbl:AUC}, we have the following observations. First, the neural models generally performed better than the Bayesian inference model (i.e. BKT). This is due to the power of these models in learning complex student learning patterns without the need to oversimplify the problem's assumption to keep it within the tractable computation limits as the case in BKT. Second, the memory-augmented models DKVMN and SKVMN performed better than the DKT model that does not utilise an external memory structure. This has empirically verified the effectiveness of external memory structures in storing past learning experiences of students, as well as  facilitating the access of relevant information to enhance the prediction performance. Third, the use of sequential dependencies among exercises in our SKVMN model enhanced the prediction accuracy in comparison to the DKVMN model which primarily considers the latest observed exercise.

\subsection{Clustering Questions}
To provide insights on how our proposed model SKVMN can correlate questions to their latent concepts, In Figure \ref{fig:clusters}.(a)-\ref{fig:clusters}.(b), we present the clustering results of questions based on their correlated concepts in the dataset ASSISTments2009, generated by using DKVMN and SKVMN, respectively. This dataset was selected for two reasons. First, it has a reasonable number of questions (i.e., 110), enabling the visualization of clusters to be readable. Second, each question in this dataset is provided with a description as depicted in Table \ref{tab:concepts}, which is useful for validating how well the model discovers correlations between questions and their latent concepts.

As shown in Figure \ref{fig:clusters}, both DKVMN and SKVMN  discover that there are 10 latent concepts relating to the 110 questions in the dataset ASSISTments2009, where all questions in one cluster relate to common latent concepts and are labelled using the same color. It can be noticed that 
SKVMN performs significantly better than DKVMN since the overlapping between different clusters in SKVMN is smaller than in DKVMN. For example, in Figure \ref{fig:clusters}.(a),  question 105 is about curve slop, which is close to the cluster for geometric concepts in brown colour, while it is placed in the cluster for equation system concepts in blue colour. Similarly, other overlaps can be observed such as questions 38, 73, and 26. This indicates that the effectiveness of SKVMN in discovering latent concepts as well as discovering questions that relate to these latent concepts. 

We can further verify the effectiveness of SKVMN in discovering latent concepts for questions using the question descriptions in Table \ref{tab:concepts}. For example, in Figure \ref{fig:clusters}.(b), the questions $13$, $19$, and $30$ fall in the same cluster in pink (top right corner). Their provided descriptions are ``\textit{Equivalent Fractions}'',``\textit{Multiplication Fractions}'', and ``\textit{Ordering Fractions}'', respectively, which are all relevant to fractions concepts. Similarly, the questions $1$, $81$, and $92$ have the descriptions ``\textit{Area Trapezoid}'', ``\textit{Area Rectangle}'', and ``\textit{Rotations}'', respectively. These questions fall in the same cluster in light blue (bottom right corner) because they are about geometric concepts, such as area functions and transformations. 

Note that, SKVMN depends on identity vectors to aggregate questions with common concepts together. While the DKVMN depends on the attention vectors to perform this aggregation. 

\subsection{Evolution of Knowledge States}\label{subsec:evolution-knowledge-state}

As previously discussed in Section \ref{sec:intro}, the knowledge states of a student may evolve over time, through learning from a sequence of exercises. In order to illustrate this evolution process, Figure  \ref{fig:heatmap} shows a student's knowledge states over a sequence of 50 exercises from the ASSISTments2009 dataset. At each time step, a knowledge state consists of the concept states of five concepts $\mathrm{\{c^1, \dots, c^5\}}$, which are stored in the value matrix of a key-value memory augmented with DKVMN or SKVMN. Figure \ref{fig:heatmap}.(a) shows this student's knowledge states captured by DKVMN, while Figure \ref{fig:heatmap}.(b) shows this student's knowledge states captured by SKVMN. 

In SKVMN, relevant questions are identified as shown in Table \ref{tab:concepts}. Comparing Figure \ref{fig:heatmap}.(a) and  Figure \ref{fig:heatmap}.(b), it can be visually noticed that SKVMN has smoother updates to the concept states in the value matrix than DKVMN. For example, considering the questions 23, 33 and 61, the student answered the first two questions incorrectly and the last one correctly, which result in a sudden update to the value of $\mathrm{c^{3}}$ (i.e., the concept state of $\mathrm{c^3}$) in the value matrix of DKVMN but a smoother update to the concept state of $\mathrm{c^3}$ in the value matrix of SKVMN. Another example is the questions 70 and 88 that correlate to same latent concepts, the student answered the first one incorrectly and the second one correctly, which resulted in a significant update to the concept state of $\mathrm{c^3}$ in the DKVMN's memory around indices 18, 19 and 20, while SKVMN's concept state of $\mathrm{c^3}$ decreased in a smoother manner. This means that SKVMN considers the past performance of the student in relevance to this concept. At its core, these differences in capturing concept states are due to the fact that DKVMN's write process only takes the question and the answer to calculate the erase and add vectors, so that the knowledge state of DKVMN is biased to the most recently observed question. SKVNM has resolved this issue by taking into account the summary vector (i.e., current knowledge state and the level of difficulty of the current question) for the write process.

%% file: section_RelatedWork.tex
\section{Related Work}
\label{sec:rw}

In this section, we provide a brief review of related research work.

One of the early attempts for developing a KT model was introduced by Corbett and Anderson \citep{Corbett1994}. Their KT model, called {Bayesian Knowledge Tracing} (BKT), assumed a knowledge state to be a binary random variable (i.e. know or do not know) and followed a Bayesian inference approach to estimate the values of knowledge states. 
However, BKT has limitations in modelling dynamics between different concepts due to its oversimplified representation to make the Bayesian inference tractable. Baker et al. \citep{Baker2008} extended BKT by introducing an additional layer to the Bayesian inference to represent the contextual information. While their model achieved better results, it was still considered only the latest observation as a first-order Markov chain. Several attempts have been made to extend BKT by individualizing the prior distribution of Bayesian inference parameters \citep{Pardos_2010,Yudelson_2013} so as to customize the model for each individual student. These individualization techniques were proved to reduce prediction errors of the original BKT model. Pardos and Heffernan \citep{Pardos_2011} introduced the use of auxiliary information to the Bayesian inference process, such as item difficulty, and showed that it can further enhance the prediction accuracy.

With the rise of deep learning models \citep{lecun2015deep} and their achieved breakthroughs in sequence modelling \cite{SCHMIDHUBER201585}, such as natural language processing \cite{NLP1,NLP2}, video recognition \cite{Video1,video2}, and signal processing \cite{Signal}, recent studies adopted deep learning models to address the KT problem. Piech et al. \cite{DKT2015_5654} proposed the deep knowledge tracing (DKT) model which uses a recurrent neural networks (RNN) \citep{Mandic_2001} to model dynamics in a past exercise sequence and predicts answers for new questions. DKT resolved the limitations of Bayesian inference approaches as RNNs optimization through backpropagation is tractable. Despite this advance, DKT assumed only one hidden state variable for representing a student's knowledge state, which is an unrealistic assumption for real-world scenarios as a student's knowledge can significantly vary across different learning concepts. To address this limitation, Zhang et al. \citep{DKVMN17} proposed a model called Dynamic Key-Value Memory Networks (DKVMN), which followed the concepts of Memory-Augmented Neural Networks (MANN) \citep{pmlr-v48-santoro16,Graves2016}. MANN aim at mimicking the human's brain functionality which combines neural spiking for computation with memory for storing past experiences \citep{gallistel2011memory}. Inspired by MANN, DKVMN is augmented with two auxiliary memory structures: the key matrix and the value matrix. The former is used to keep the concepts underlying exercises, while the later one is used to store a knowledge state across these concepts. Results showed that DKVMN outperformed BKT and DKT on standard KT benchmarks, and therefore it is considered the state-of-the-art KT models. However, DKVMN only considers the latest exercise embedding when updating the value matrix, resulting in biased knowledge states that ignore past learning experience. As an example, if we have three related exercises in a sequence, two being answered correctly and the latest being answered incorrectly, DKVMN would be biased to the latest one and update the knowledge state with knowledge loss abruptly. In addition to this, DKVMN has no model capacity to capture long dependencies in an exercise sequence. This assumes a first-order Markov chain to represent a past exercise sequence, which is not satisfactory in many scenarios. Our proposed KT model has addressed the limitations from both DKT and DKVMN. 

 In our proposed KT model, we developed a modified LSTM, called Hop-LSTM, for sequence modelling. Current recurrent neural network models (RNNs) and their variants, such as LSTMs \cite{lstm97}, bi-directional RNNs \cite{BIRNN}, or other gated RNNs \cite{GRNN}, provide the capacity to effectively ingest dependencies in sequential data. However, when sequences are long, it is still difficult to capture long term dependencies. One way to alleviate this issue is to only update a fraction of hidden states based on the current hidden state and input \cite{jernite2016variable}. For example, Yu, Lee and Le \cite{NLP2} proposed a LSTM model that can jump ahead in a sequence to avoid irrelevant words. The jump decision was controlled by a policy gradient reinforcement learning algorithm that works as an active learning technique to sample only important words for the model. Campos et al. \cite{campos2018skip} proposed a model by augmenting the standard RNN with a binary state update gate function which is responsible for deciding whether to update the hidden state or not based on the number of previous updates performed and a loss term that balances the number of updates (i.e. learning speed) with achieved accuracy. Different from these models, we developed Hop-LSTM in relation to a Triangular layer so that only the hidden states of LSTM cells for relevant exercises are connected. This allows our KT model to identify relevant skills and prior background from the past learning activities (e.g., exercise sequences) for improved prediction accuracy.

%% file: section_Conclusion.tex
\section{Conclusions}
\label{sec:conc}
 In this paper, we introduced a novel model called Sequential Key-Value Memory Networks (SKVMN) for knowledge tracing. SKVMN aimed at overcoming the limitations of the existing KT models. It is augmented with a key-value memory at the memory layer and a modified LSTM, called Hop-LSTM, at the sequence layer. The experimental results showed that our proposed model outperformed the state-of-the-art models over all datasets.
Future work will consider techniques to automatically tune hyper-parameters for Knowledge Tracing models.

\begin{acks}
This research is supported by an Australian government higher education scholarship, ANU Vice-Chancellor's Teaching Enhancement Grant, as well as NVIDIA for the generous GPU support.
\end{acks}

%% file: main.bbl

\begin{thebibliography}{00}


\ifx \showCODEN    \undefined \def \showCODEN     #1{\unskip}     \fi
\ifx \showDOI      \undefined \def \showDOI       #1{#1}\fi
\ifx \showISBNx    \undefined \def \showISBNx     #1{\unskip}     \fi
\ifx \showISBNxiii \undefined \def \showISBNxiii  #1{\unskip}     \fi
\ifx \showISSN     \undefined \def \showISSN      #1{\unskip}     \fi
\ifx \showLCCN     \undefined \def \showLCCN      #1{\unskip}     \fi
\ifx \shownote     \undefined \def \shownote      #1{#1}          \fi
\ifx \showarticletitle \undefined \def \showarticletitle #1{#1}   \fi
\ifx \showURL      \undefined \def \showURL       {\relax}        \fi
\providecommand\bibfield[2]{#2}
\providecommand\bibinfo[2]{#2}
\providecommand\natexlab[1]{#1}
\providecommand\showeprint[2][]{arXiv:#2}

\bibitem[\protect\citeauthoryear{Baker, Corbett, and Aleven}{Baker
  et~al\mbox{.}}{2008}]%
        {Baker2008}
\bibfield{author}{\bibinfo{person}{Ryan~S. Baker}, \bibinfo{person}{Albert~T.
  Corbett}, {and} \bibinfo{person}{Vincent Aleven}.}
  \bibinfo{year}{2008}\natexlab{}.
\newblock \showarticletitle{More Accurate Student Modeling Through Contextual
  Estimation of Slip and Guess Probabilities in Bayesian Knowledge Tracing}. In
  \bibinfo{booktitle}{{\em Proceedings of the 9th International Conference on
  Intelligent Tutoring Systems}} {\em (\bibinfo{series}{ITS})}.
  \bibinfo{address}{Berlin, Heidelberg}, \bibinfo{pages}{406--415}.
\newblock
\showISBNx{978-3-540-69130-3}


\bibitem[\protect\citeauthoryear{Bengio}{Bengio}{2012}]%
        {Bengio2012}
\bibfield{author}{\bibinfo{person}{Yoshua Bengio}.}
  \bibinfo{year}{2012}\natexlab{}.
\newblock \showarticletitle{Practical recommendations for gradient-based
  training of deep architectures}.
\newblock In \bibinfo{booktitle}{{\em Neural networks: Tricks of the trade:
  Second Edition}}. \bibinfo{address}{Berlin, Heidelberg},
  \bibinfo{pages}{437--478}.
\newblock


\bibitem[\protect\citeauthoryear{Bottou}{Bottou}{2012}]%
        {Bottou2012}
\bibfield{author}{\bibinfo{person}{L{\'e}on Bottou}.}
  \bibinfo{year}{2012}\natexlab{}.
\newblock \showarticletitle{Stochastic gradient descent tricks}.
\newblock In \bibinfo{booktitle}{{\em Neural networks: Tricks of the trade:
  Second Edition}}. \bibinfo{address}{Berlin, Heidelberg},
  \bibinfo{pages}{421--436}.
\newblock


\bibitem[\protect\citeauthoryear{Campos, Jou, {Gir{\'{o}} i Nieto}, Torres, and
  Chang}{Campos et~al\mbox{.}}{2018}]%
        {campos2018skip}
\bibfield{author}{\bibinfo{person}{V{\'{\i}}ctor Campos},
  \bibinfo{person}{Brendan Jou}, \bibinfo{person}{Xavier {Gir{\'{o}} i Nieto}},
  \bibinfo{person}{Jordi Torres}, {and} \bibinfo{person}{Shih{-}Fu Chang}.}
  \bibinfo{year}{2018}\natexlab{}.
\newblock \showarticletitle{Skip {RNN}: Learning to Skip State Updates in
  Recurrent Neural Networks}. In \bibinfo{booktitle}{{\em 6th International
  Conference on Learning Representations, (ICLR), Vancouver, BC, Canada, April
  30 - May 3, 2018, Conference Track Proceedings}}.
\newblock


\bibitem[\protect\citeauthoryear{Chang, Hsu, and Chen}{Chang
  et~al\mbox{.}}{2015}]%
        {chang2015modeling}
\bibfield{author}{\bibinfo{person}{Haw-Shiuan Chang},
  \bibinfo{person}{Hwai-Jung Hsu}, {and} \bibinfo{person}{Kuan-Ta Chen}.}
  \bibinfo{year}{2015}\natexlab{}.
\newblock \showarticletitle{Modeling Exercise Relationships in E-Learning: A
  Unified Approach.}. In \bibinfo{booktitle}{{\em Proceedings of the 8th
  International Conference on Educational Data Mining, (EDM), Madrid, Spain,
  June 26-29, 2015}}. \bibinfo{pages}{532--535}.
\newblock


\bibitem[\protect\citeauthoryear{Corbett and Anderson}{Corbett and
  Anderson}{1994}]%
        {Corbett1994}
\bibfield{author}{\bibinfo{person}{Albert~T. Corbett} {and}
  \bibinfo{person}{John~R. Anderson}.} \bibinfo{year}{1994}\natexlab{}.
\newblock \showarticletitle{Knowledge tracing: Modeling the acquisition of
  procedural knowledge}.
\newblock \bibinfo{journal}{{\em User Modeling and User-Adapted Interaction\/}}
  \bibinfo{volume}{4}, \bibinfo{number}{4} (\bibinfo{date}{01 Dec}
  \bibinfo{year}{1994}), \bibinfo{pages}{253--278}.
\newblock
\showISSN{1573-1391}


\bibitem[\protect\citeauthoryear{Donahue, Anne~Hendricks, Guadarrama, Rohrbach,
  Venugopalan, Saenko, and Darrell}{Donahue et~al\mbox{.}}{2015}]%
        {Video1}
\bibfield{author}{\bibinfo{person}{Jeffrey Donahue}, \bibinfo{person}{Lisa
  Anne~Hendricks}, \bibinfo{person}{Sergio Guadarrama}, \bibinfo{person}{Marcus
  Rohrbach}, \bibinfo{person}{Subhashini Venugopalan}, \bibinfo{person}{Kate
  Saenko}, {and} \bibinfo{person}{Trevor Darrell}.}
  \bibinfo{year}{2015}\natexlab{}.
\newblock \showarticletitle{Long-Term Recurrent Convolutional Networks for
  Visual Recognition and Description}. In \bibinfo{booktitle}{{\em The IEEE
  Conference on Computer Vision and Pattern Recognition, (CVPR) , Boston, MA,
  USA, June 7-12, 2015}}. \bibinfo{pages}{2625--2634}.
\newblock


\bibitem[\protect\citeauthoryear{Gallistel and King}{Gallistel and
  King}{2011}]%
        {gallistel2011memory}
\bibfield{author}{\bibinfo{person}{Charles~R Gallistel} {and}
  \bibinfo{person}{Adam~Philip King}.} \bibinfo{year}{2011}\natexlab{}.
\newblock \bibinfo{booktitle}{{\em Memory and the computational brain: Why
  cognitive science will transform neuroscience}}. Vol.~\bibinfo{volume}{6}.
\newblock \bibinfo{publisher}{John Wiley \& Sons}.
\newblock


\bibitem[\protect\citeauthoryear{Glorot and Bengio}{Glorot and Bengio}{2010}]%
        {glorot2010}
\bibfield{author}{\bibinfo{person}{Xavier Glorot} {and} \bibinfo{person}{Yoshua
  Bengio}.} \bibinfo{year}{2010}\natexlab{}.
\newblock \showarticletitle{Understanding the difficulty of training deep
  feedforward neural networks}. In \bibinfo{booktitle}{{\em In Proceedings of
  the International Conference on Artificial Intelligence and Statistics
  (AISTATS). Society for Artificial Intelligence and Statistics}}.
  \bibinfo{address}{Chia Laguna Resort, Sardinia, Italy},
  \bibinfo{pages}{249--256}.
\newblock


\bibitem[\protect\citeauthoryear{Graves, Jaitly, and Mohamed}{Graves
  et~al\mbox{.}}{2013a}]%
        {NLP1}
\bibfield{author}{\bibinfo{person}{A. Graves}, \bibinfo{person}{N. Jaitly},
  {and} \bibinfo{person}{A. Mohamed}.} \bibinfo{year}{2013}\natexlab{a}.
\newblock \showarticletitle{Hybrid speech recognition with Deep Bidirectional
  LSTM}. In \bibinfo{booktitle}{{\em 2013 {IEEE} Workshop on Automatic Speech
  Recognition and Understanding, Olomouc, Czech Republic, December 8-12,
  2013}}. \bibinfo{pages}{273--278}.
\newblock


\bibitem[\protect\citeauthoryear{Graves, Mohamed, and Hinton}{Graves
  et~al\mbox{.}}{2013b}]%
        {Graves13}
\bibfield{author}{\bibinfo{person}{A. Graves}, \bibinfo{person}{A. Mohamed},
  {and} \bibinfo{person}{G. Hinton}.} \bibinfo{year}{2013}\natexlab{b}.
\newblock \showarticletitle{Speech recognition with deep recurrent neural
  networks}. In \bibinfo{booktitle}{{\em 2013 IEEE International Conference on
  Acoustics, Speech and Signal Processing}}. \bibinfo{address}{Vancouver, BC,
  Canada}, \bibinfo{pages}{6645--6649}.
\newblock
\showISSN{1520-6149}


\bibitem[\protect\citeauthoryear{Graves, Wayne, Reynolds, Harley, Danihelka,
  Grabska-Barwinska, Colmenarejo, Grefenstette, Ramalho, Agapiou, Badia,
  Hermann, Zwols, Ostrovski, Cain, King, Summerfield, Blunsom, Kavukcuoglu, and
  Hassabis}{Graves et~al\mbox{.}}{2016}]%
        {Graves2016}
\bibfield{author}{\bibinfo{person}{Alex Graves}, \bibinfo{person}{Greg Wayne},
  \bibinfo{person}{Malcolm Reynolds}, \bibinfo{person}{Tim Harley},
  \bibinfo{person}{Ivo Danihelka}, \bibinfo{person}{Agnieszka
  Grabska-Barwinska}, \bibinfo{person}{Sergio~G{\'o}mez Colmenarejo},
  \bibinfo{person}{Edward Grefenstette}, \bibinfo{person}{Tiago Ramalho},
  \bibinfo{person}{John Agapiou}, \bibinfo{person}{Adri{\`a}~Puigdom{\`e}nech
  Badia}, \bibinfo{person}{Karl~Moritz Hermann}, \bibinfo{person}{Yori Zwols},
  \bibinfo{person}{Georg Ostrovski}, \bibinfo{person}{Adam Cain},
  \bibinfo{person}{Helen King}, \bibinfo{person}{Christopher Summerfield},
  \bibinfo{person}{Phil Blunsom}, \bibinfo{person}{Koray Kavukcuoglu}, {and}
  \bibinfo{person}{Demis Hassabis}.} \bibinfo{year}{2016}\natexlab{}.
\newblock \showarticletitle{Hybrid computing using a neural network with
  dynamic external memory}.
\newblock \bibinfo{journal}{{\em Nature\/}}  \bibinfo{volume}{538}
  (\bibinfo{date}{12 Oct} \bibinfo{year}{2016}), \bibinfo{pages}{471}.
\newblock


\bibitem[\protect\citeauthoryear{Hochreiter and Schmidhuber}{Hochreiter and
  Schmidhuber}{1997}]%
        {lstm97}
\bibfield{author}{\bibinfo{person}{Sepp Hochreiter} {and}
  \bibinfo{person}{J\"{u}rgen Schmidhuber}.} \bibinfo{year}{1997}\natexlab{}.
\newblock \showarticletitle{Long Short-Term Memory}.
\newblock \bibinfo{journal}{{\em Neural Comput.\/}} \bibinfo{volume}{9},
  \bibinfo{number}{8} (\bibinfo{date}{Nov.} \bibinfo{year}{1997}),
  \bibinfo{pages}{1735--1780}.
\newblock


\bibitem[\protect\citeauthoryear{Jernite, Grave, Joulin, and Mikolov}{Jernite
  et~al\mbox{.}}{2017}]%
        {jernite2016variable}
\bibfield{author}{\bibinfo{person}{Yacine Jernite}, \bibinfo{person}{Edouard
  Grave}, \bibinfo{person}{Armand Joulin}, {and} \bibinfo{person}{Tomas
  Mikolov}.} \bibinfo{year}{2017}\natexlab{}.
\newblock \showarticletitle{Variable computation in recurrent neural networks}.
  In \bibinfo{booktitle}{{\em 5th International Conference on Learning
  Representations, (ICLR), Toulon, France, April 24-26, 2017, Conference Track
  Proceedings}}.
\newblock


\bibitem[\protect\citeauthoryear{Khajah, Lindsey, and Mozer}{Khajah
  et~al\mbox{.}}{2016}]%
        {khajah2016deep}
\bibfield{author}{\bibinfo{person}{Mohammad Khajah}, \bibinfo{person}{Robert~V
  Lindsey}, {and} \bibinfo{person}{Michael~C Mozer}.}
  \bibinfo{year}{2016}\natexlab{}.
\newblock \showarticletitle{How Deep is Knowledge Tracing?}. In
  \bibinfo{booktitle}{{\em Proceedings of the 9th International Conference on
  Educational Data Mining, (EDM), Raleigh, North Carolina, USA, June 29 - July
  2, 2016}}.
\newblock


\bibitem[\protect\citeauthoryear{{Kingma} and {Ba}}{{Kingma} and {Ba}}{2015}]%
        {Adam2015}
\bibfield{author}{\bibinfo{person}{Diederik~P. {Kingma}} {and}
  \bibinfo{person}{Jimmy~Lei {Ba}}.} \bibinfo{year}{2015}\natexlab{}.
\newblock \showarticletitle{Adam: A Method for Stochastic Optimization}. In
  \bibinfo{booktitle}{{\em international conference on learning
  representations}} {\em (\bibinfo{series}{ICLR})}.
\newblock


\bibitem[\protect\citeauthoryear{Klir and Yuan}{Klir and Yuan}{1995}]%
        {klir1995fuzzy}
\bibfield{author}{\bibinfo{person}{George~J. Klir} {and} \bibinfo{person}{Bo
  Yuan}.} \bibinfo{year}{1995}\natexlab{}.
\newblock \showarticletitle{Fuzzy Sets and Fuzzy Logic: Theory and
  Applications}.
\newblock \bibinfo{publisher}{Prentice-Hall, Inc.}, \bibinfo{address}{Upper
  Saddle River, NJ, USA}.
\newblock
\showISBNx{0-13-101171-5}


\bibitem[\protect\citeauthoryear{Kusupati, Singh, Bhatia, Kumar, Jain, and
  Varma}{Kusupati et~al\mbox{.}}{2018}]%
        {GRNN}
\bibfield{author}{\bibinfo{person}{Aditya Kusupati}, \bibinfo{person}{Manish
  Singh}, \bibinfo{person}{Kush Bhatia}, \bibinfo{person}{Ashish Kumar},
  \bibinfo{person}{Prateek Jain}, {and} \bibinfo{person}{Manik Varma}.}
  \bibinfo{year}{2018}\natexlab{}.
\newblock \showarticletitle{FastGRNN: A Fast, Accurate, Stable and Tiny
  Kilobyte Sized Gated Recurrent Neural Network}. In \bibinfo{booktitle}{{\em
  Advances in Neural Information Processing Systems 31: Annual Conference on
  Neural Information Processing Systems, (NeurIPS), 3-8 December 2018,
  Montr{\'{e}}al, Canada.}} \bibinfo{pages}{9017--9028}.
\newblock


\bibitem[\protect\citeauthoryear{LeCun, Bengio, and Hinton}{LeCun
  et~al\mbox{.}}{2015}]%
        {lecun2015deep}
\bibfield{author}{\bibinfo{person}{Yann LeCun}, \bibinfo{person}{Yoshua
  Bengio}, {and} \bibinfo{person}{Geoffrey Hinton}.}
  \bibinfo{year}{2015}\natexlab{}.
\newblock \showarticletitle{Deep learning}.
\newblock \bibinfo{journal}{{\em nature\/}} \bibinfo{volume}{521},
  \bibinfo{number}{7553} (\bibinfo{year}{2015}), \bibinfo{pages}{436}.
\newblock


\bibitem[\protect\citeauthoryear{Ling, Huang, and Zhang}{Ling
  et~al\mbox{.}}{2003}]%
        {ling2003auc}
\bibfield{author}{\bibinfo{person}{Charles~X. Ling}, \bibinfo{person}{Jin
  Huang}, {and} \bibinfo{person}{Harry Zhang}.}
  \bibinfo{year}{2003}\natexlab{}.
\newblock \showarticletitle{AUC: A Statistically Consistent and More
  Discriminating Measure Than Accuracy}. In \bibinfo{booktitle}{{\em
  Proceedings of the 18th International Joint Conference on Artificial
  Intelligence}} {\em (\bibinfo{series}{IJCAI})}. \bibinfo{address}{San
  Francisco, CA, USA}, \bibinfo{pages}{519--524}.
\newblock


\bibitem[\protect\citeauthoryear{Liu, Shahroudy, Xu, and Wang}{Liu
  et~al\mbox{.}}{2016}]%
        {video2}
\bibfield{author}{\bibinfo{person}{Jun Liu}, \bibinfo{person}{Amir Shahroudy},
  \bibinfo{person}{Dong Xu}, {and} \bibinfo{person}{Gang Wang}.}
  \bibinfo{year}{2016}\natexlab{}.
\newblock \showarticletitle{Spatio-Temporal LSTM with Trust Gates for 3D Human
  Action Recognition}. In \bibinfo{booktitle}{{\em 14th European Conference on
  Computer Vision, (ECCV) , Amsterdam, The Netherlands, October 11-14, 2016,
  Proceedings, Part {III}}}. \bibinfo{address}{Cham},
  \bibinfo{pages}{816--833}.
\newblock


\bibitem[\protect\citeauthoryear{Mandic and Chambers}{Mandic and
  Chambers}{2001}]%
        {Mandic_2001}
\bibfield{author}{\bibinfo{person}{Danilo~P. Mandic} {and}
  \bibinfo{person}{Jonathon Chambers}.} \bibinfo{year}{2001}\natexlab{}.
\newblock \bibinfo{booktitle}{{\em Recurrent Neural Networks for Prediction:
  Learning Algorithms,Architectures and Stability}}.
\newblock \bibinfo{publisher}{John Wiley \& Sons, Inc.}, \bibinfo{address}{New
  York, NY, USA}.
\newblock
\showISBNx{0471495174}


\bibitem[\protect\citeauthoryear{Pardos and Heffernan}{Pardos and
  Heffernan}{2010}]%
        {Pardos_2010}
\bibfield{author}{\bibinfo{person}{Zachary~A. Pardos} {and}
  \bibinfo{person}{Neil~T. Heffernan}.} \bibinfo{year}{2010}\natexlab{}.
\newblock \showarticletitle{Modeling Individualization in a Bayesian Networks
  Implementation of Knowledge Tracing}. In \bibinfo{booktitle}{{\em Proceedings
  of the 18th International Conference on User Modeling, Adaptation, and
  Personalization}} {\em (\bibinfo{series}{UMAP})}. \bibinfo{address}{Berlin,
  Heidelberg}, \bibinfo{pages}{255--266}.
\newblock
\showISBNx{3-642-13469-6, 978-3-642-13469-2}


\bibitem[\protect\citeauthoryear{Pardos and Heffernan}{Pardos and
  Heffernan}{2011}]%
        {Pardos_2011}
\bibfield{author}{\bibinfo{person}{Zachary~A. Pardos} {and}
  \bibinfo{person}{Neil~T. Heffernan}.} \bibinfo{year}{2011}\natexlab{}.
\newblock \showarticletitle{KT-IDEM: Introducing Item Difficulty to the
  Knowledge Tracing Model}. In \bibinfo{booktitle}{{\em Proceedings of the 19th
  International Conference on User Modeling, Adaption, and Personalization}}
  {\em (\bibinfo{series}{UMAP})}. \bibinfo{address}{Berlin, Heidelberg},
  \bibinfo{pages}{243--254}.
\newblock
\showISBNx{978-3-642-22361-7}


\bibitem[\protect\citeauthoryear{Pascanu, Mikolov, and Bengio}{Pascanu
  et~al\mbox{.}}{2013}]%
        {pascanu2013difficulty}
\bibfield{author}{\bibinfo{person}{Razvan Pascanu}, \bibinfo{person}{Tomas
  Mikolov}, {and} \bibinfo{person}{Yoshua Bengio}.}
  \bibinfo{year}{2013}\natexlab{}.
\newblock \showarticletitle{On the Difficulty of Training Recurrent Neural
  Networks}. In \bibinfo{booktitle}{{\em Proceedings of the 30th International
  Conference on International Conference on Machine Learning}} {\em
  (\bibinfo{series}{ICML})}. \bibinfo{address}{Atlanta, Georgia, USA},
  \bibinfo{pages}{III--1310--III--1318}.
\newblock


\bibitem[\protect\citeauthoryear{Piech, Bassen, Huang, Ganguli, Sahami, Guibas,
  and Sohl-Dickstein}{Piech et~al\mbox{.}}{2015}]%
        {DKT2015_5654}
\bibfield{author}{\bibinfo{person}{Chris Piech}, \bibinfo{person}{Jonathan
  Bassen}, \bibinfo{person}{Jonathan Huang}, \bibinfo{person}{Surya Ganguli},
  \bibinfo{person}{Mehran Sahami}, \bibinfo{person}{Leonidas Guibas}, {and}
  \bibinfo{person}{Jascha Sohl-Dickstein}.} \bibinfo{year}{2015}\natexlab{}.
\newblock \showarticletitle{Deep Knowledge Tracing}. In
  \bibinfo{booktitle}{{\em Advances in Neural Information Processing Systems
  28: Annual Conference on Neural Information Processing Systems 2015, December
  7-12, 2015, Montreal, Quebec, Canada}} {\em (\bibinfo{series}{NeurIPS})}.
  \bibinfo{address}{Cambridge, MA, USA}, \bibinfo{pages}{505--513}.
\newblock


\bibitem[\protect\citeauthoryear{Piech}{Piech}{2016}]%
        {piech2016uncovering}
\bibfield{author}{\bibinfo{person}{Christopher~James Piech}.}
  \bibinfo{year}{2016}\natexlab{}.
\newblock {\em \bibinfo{title}{Uncovering Patterns in Student Work: Machine
  Learning to Understand Human Learning}}.
\newblock \bibinfo{thesistype}{Ph.D. Dissertation}. \bibinfo{school}{Stanford
  University}.
\newblock


\bibitem[\protect\citeauthoryear{Santoro, Bartunov, Botvinick, Wierstra, and
  Lillicrap}{Santoro et~al\mbox{.}}{2016}]%
        {pmlr-v48-santoro16}
\bibfield{author}{\bibinfo{person}{Adam Santoro}, \bibinfo{person}{Sergey
  Bartunov}, \bibinfo{person}{Matthew Botvinick}, \bibinfo{person}{Daan
  Wierstra}, {and} \bibinfo{person}{Timothy Lillicrap}.}
  \bibinfo{year}{2016}\natexlab{}.
\newblock \showarticletitle{Meta-learning with Memory-augmented Neural
  Networks}. In \bibinfo{booktitle}{{\em Proceedings of the 33nd International
  Conference on Machine Learning, (ICML), New York City, NY, USA, June 19-24,
  2016}}. \bibinfo{pages}{1842--1850}.
\newblock


\bibitem[\protect\citeauthoryear{Schmidhuber}{Schmidhuber}{2015}]%
        {SCHMIDHUBER201585}
\bibfield{author}{\bibinfo{person}{J{\"u}rgen Schmidhuber}.}
  \bibinfo{year}{2015}\natexlab{}.
\newblock \showarticletitle{Deep learning in neural networks: An overview}.
\newblock \bibinfo{journal}{{\em Neural Netw.\/}} \bibinfo{volume}{61},
  \bibinfo{number}{C} (\bibinfo{date}{Jan.} \bibinfo{year}{2015}),
  \bibinfo{pages}{85--117}.
\newblock
\showISSN{0893-6080}


\bibitem[\protect\citeauthoryear{Schuster and Paliwal}{Schuster and
  Paliwal}{1997}]%
        {BIRNN}
\bibfield{author}{\bibinfo{person}{M. Schuster} {and} \bibinfo{person}{K.~K.
  Paliwal}.} \bibinfo{year}{1997}\natexlab{}.
\newblock \showarticletitle{Bidirectional recurrent neural networks}.
\newblock \bibinfo{journal}{{\em IEEE Transactions on Signal Processing\/}}
  \bibinfo{volume}{45}, \bibinfo{number}{11} (\bibinfo{date}{Nov}
  \bibinfo{year}{1997}), \bibinfo{pages}{2673--2681}.
\newblock
\showISSN{1053-587X}


\bibitem[\protect\citeauthoryear{Sutskever, Vinyals, and Le}{Sutskever
  et~al\mbox{.}}{2014}]%
        {Sutskever_2014}
\bibfield{author}{\bibinfo{person}{Ilya Sutskever}, \bibinfo{person}{Oriol
  Vinyals}, {and} \bibinfo{person}{Quoc~V. Le}.}
  \bibinfo{year}{2014}\natexlab{}.
\newblock \showarticletitle{Sequence to Sequence Learning with Neural
  Networks}. In \bibinfo{booktitle}{{\em Advances in Neural Information
  Processing Systems 27: Annual Conference on Neural Information Processing
  Systems, December 8-13 2014, Montreal, Quebec, Canada}} {\em
  (\bibinfo{series}{NeurIPS})}. \bibinfo{address}{Cambridge, MA, USA},
  \bibinfo{pages}{3104--3112}.
\newblock


\bibitem[\protect\citeauthoryear{Villano}{Villano}{1992}]%
        {Villano92}
\bibfield{author}{\bibinfo{person}{Michael Villano}.}
  \bibinfo{year}{1992}\natexlab{}.
\newblock \showarticletitle{Probabilistic Student Models: Bayesian Belief
  Networks and Knowledge Space Theory}. In \bibinfo{booktitle}{{\em Proceedings
  of the Second International Conference on Intelligent Tutoring Systems}} {\em
  (\bibinfo{series}{ITS})}. \bibinfo{address}{London, UK, UK},
  \bibinfo{pages}{491--498}.
\newblock
\showISBNx{3-540-55606-0}


\bibitem[\protect\citeauthoryear{Wang, Jiang, Liu, Shang, and Zhang}{Wang
  et~al\mbox{.}}{2018}]%
        {Signal}
\bibfield{author}{\bibinfo{person}{P. Wang}, \bibinfo{person}{A. Jiang},
  \bibinfo{person}{X. Liu}, \bibinfo{person}{J. Shang}, {and}
  \bibinfo{person}{L. Zhang}.} \bibinfo{year}{2018}\natexlab{}.
\newblock \showarticletitle{LSTM-Based EEG Classification in Motor Imagery
  Tasks}.
\newblock \bibinfo{journal}{{\em IEEE Transactions on Neural Systems and
  Rehabilitation Engineering\/}} \bibinfo{volume}{26}, \bibinfo{number}{11}
  (\bibinfo{date}{Nov} \bibinfo{year}{2018}), \bibinfo{pages}{2086--2095}.
\newblock
\showISSN{1534-4320}


\bibitem[\protect\citeauthoryear{{Yu}, {Lee}, and {Le}}{{Yu}
  et~al\mbox{.}}{2017}]%
        {NLP2}
\bibfield{author}{\bibinfo{person}{Adams~Wei {Yu}}, \bibinfo{person}{Hongrae
  {Lee}}, {and} \bibinfo{person}{Quoc~V. {Le}}.}
  \bibinfo{year}{2017}\natexlab{}.
\newblock \showarticletitle{{Learning to Skim Text}}. In
  \bibinfo{booktitle}{{\em Proceedings of the 55th Annual Meeting of the
  Association for Computational Linguistics, (ACL), Vancouver, Canada, July 30
  - August 4, (Volume 1: Long Papers)}}. \bibinfo{pages}{1880--1890}.
\newblock


\bibitem[\protect\citeauthoryear{Yudelson, Koedinger, and Gordon}{Yudelson
  et~al\mbox{.}}{2013}]%
        {Yudelson_2013}
\bibfield{author}{\bibinfo{person}{Michael~V Yudelson},
  \bibinfo{person}{Kenneth~R Koedinger}, {and} \bibinfo{person}{Geoffrey~J
  Gordon}.} \bibinfo{year}{2013}\natexlab{}.
\newblock \showarticletitle{Individualized bayesian knowledge tracing models}.
  In \bibinfo{booktitle}{{\em Artificial Intelligence in Education - 16th
  International Conference, (AIED), Memphis, TN, USA, July 9-13, 2013.
  Proceedings}}. \bibinfo{pages}{171--180}.
\newblock


\bibitem[\protect\citeauthoryear{Zhang, Shi, King, and Yeung}{Zhang
  et~al\mbox{.}}{2017}]%
        {DKVMN17}
\bibfield{author}{\bibinfo{person}{Jiani Zhang}, \bibinfo{person}{Xingjian
  Shi}, \bibinfo{person}{Irwin King}, {and} \bibinfo{person}{Dit-Yan Yeung}.}
  \bibinfo{year}{2017}\natexlab{}.
\newblock \showarticletitle{Dynamic Key-Value Memory Networks for Knowledge
  Tracing}. In \bibinfo{booktitle}{{\em Proceedings of the 26th International
  Conference on World Wide Web}} {\em (\bibinfo{series}{WWW})}.
  \bibinfo{address}{Republic and Canton of Geneva, Switzerland},
  \bibinfo{pages}{765--774}.
\newblock
\showISBNx{978-1-4503-4913-0}


\end{thebibliography}
